\documentclass[11pt]{article}

\usepackage[utf8]{inputenc}
\usepackage[T1]{fontenc}
\usepackage{lmodern}
\usepackage[margin=1in]{geometry}
\usepackage{amsmath,amssymb}
\usepackage{bm}
\usepackage{graphicx}
\usepackage{booktabs}
\usepackage{multirow}
\usepackage{caption}
\usepackage[american]{babel}
\usepackage{csquotes}
\usepackage{authblk}
\usepackage{hyperref}
\hypersetup{
  colorlinks=false,
  linkbordercolor={0 1 0},
  citebordercolor={0 1 0},
  urlbordercolor={0 1 0},
}

\usepackage[style=apa,backend=biber,natbib=true]{biblatex}
\DeclareLanguageMapping{american}{american-apa}
\graphicspath{{figures/}}
\title{\bfseries Estimating Uncertainty in Classifier Performance with
Applications to Large Language Models and Nested Data}

\author[1]{Kylie Anglin, PhD\thanks{Kylie Anglin is the corresponding author
and can be reached at: Email: Kylie.Anglin@uconn.edu; Phone: 860-486-0181.}}
\affil[1]{Department of Educational Psychology, Neag School of Education,
University of Connecticut}
\date{}

\begin{document}
\maketitle

\begin{abstract}
\noindent Researchers increasingly use text classification---supervised models or large language models---to measure constructs from natural language, providing metrics such as recall and precision as evidence of their validity. Yet, though these metrics are point estimates subject to sampling variation, measures of uncertainty are inconsistently reported alongside them. Further, when they are reported, they are often estimated with methods that are not appropriate when relevant labelled datasets are small or performance is high. To increase and improve confidence interval reporting in the field, this paper evaluates confidence interval methods for performance metrics under conditions typical of social science text classification: small to moderate sample sizes, infrequent constructs, and texts nested within individuals. Across simulations, default methods such as the Wald interval and the basic percentile bootstrap are the least accurate, with coverage sometimes far below the nominal 95\% level. Accuracy is improved with the use of Agresti-Coull, Wilson, Clopper-Pearson, and a novel pseudo-count regularized bootstrap (which is particularly relevant to the calculation of F1). When texts are nested within individuals, we demonstrate that adjustment for both effective N and the appropriate degrees of freedom is necessary for producing accurate analytic intervals. Among bootstrap intervals, the hierarchical bootstrap is more accurate than the cluster bootstrap when individuals produce a moderate number of texts but overly conservative when individuals produce only a few. By providing guidance to the field on appropriate interval estimation, we aim to improve the transparency of machine learning applications, and to encourage greater attention to the validation sample size at the design stage.
\end{abstract}

\section*{Introduction}

Text classification---the automated sorting of text into categories, whether through the training of supervised machine learning models or through the employment of pre-trained large language models (LLMs)---allows researchers to extract measures from natural language data, in large amounts, and at a relatively limited expense. However, the resulting classifications are nearly inevitably accompanied by error; a subset of texts will likely be classified into categories for which a human expert would disagree. To assess the extent of this error, researchers commonly report evaluation metrics (like precision and recall) that have been estimated within a sample of hand-classified texts \citep{grimmer_text_2022, james_introduction_2013}. These metrics give the reader, and potentially downstream users, information on the degree of alignment between the machine classifications and expert classifications \citep{anglin_addressing_2024}, and on frequency and types of error \citep{mitchell_model_2019}. Thus, the magnitude of a given performance metric is essential for assessing the degree of trust which should be placed in the resulting inferences; while we may trust conclusions resulting from a classifier with a recall of 0.90 (identifying 90\% of instances of a concept of interest), we will have less confidence in a classifier that only identifies 75\% of instances. Further, while a reported recall of 0.90 (or 0.80, 0.70, etc., depending on the application) might be deemed good enough to employ the classifier, some lower value would not. The magnitude of the metric matters.

Yet, reported performance metrics are nothing more than point estimates---an \emph{approximation} of some true population parameter which we would obtain if we had access to an infinite sample of machine-human classification pairs. By chance, our hypothetical classifier might have an estimated recall of 0.90 in the random sample of hand-labelled texts but a true recall much lower (or higher) if we were to evaluate the population of texts. This kind of statistical uncertainty is the reason that researchers report treatment effect estimates, and other key results, alongside estimates of standard errors and/or confidence intervals. This is also the reason that researchers engage in prospective power analyses when planning randomized experiments \citep{cohen_statistical_1988, hedges_statistical_2000}; they want to ensure their sample size is large enough to reliably identify effects.

Yet, consideration of statistical uncertainty surrounding performance metrics is limited in the social sciences, including in psychology. In an informal PsycINFO search of peer reviewed articles using large language models (LLMs) and reporting performance metrics---identified using searches ``AB (large language model) AND (precision AND recall)'' and ``AB (large language model) AND (F1)''---fewer than half of the identified articles reported any measure of uncertainty surrounding those metrics (either confidence intervals or standard errors). Further, where confidence intervals for metrics are reported, they are often calculated using methods which are often inaccurate for high proportions (as is likely when reporting metrics resulting from an LLM-based classifier) and/or for nested data (e.g., when the same author produces multiple texts).

This article assesses best-practice in the estimation and reporting of performance metric uncertainty. In doing so, we address three challenges that often arise with text classification: small effective sample sizes (due to rare constructs or challenges in recruiting), performance metrics that are near 1 (due to the capabilities of LLMs), and non-independence (due to clustering of texts within authors, participants, or documents). We assess interval coverage and width through simulations under conditions typical to text classification, providing practical recommendations for confidence interval reporting. 

\section{Classifier Performance and Validity}

Text classification may be thought of as a measurement problem: there is some true---but latent---construct which is revealed within the text and which we hope our classifier can identify \citep{cronbach_construct_1955, stavropoulos_shadows_2024}. The validity of the classifier, then, depends on the extent to which the classifications reflect the construct of interest. Most commonly, this is assessed by comparing expert, human classifications to the machine classifications, under the assumption that the human classifications are closely aligned with the construct \citep{grimmer_text_2022}. Performance metrics provide an estimation of this alignment.

\subsection{Performance Metrics}

\subsubsection{Categories \texorpdfstring{$k = 2$}{k = 2}}

Texts are often classified into one of two categories, positive or negative, where the positive label usually denotes a deviation from baseline \citep{rainio_evaluation_2024}. The goal in the evaluation of binary classifiers is to assess patterns of error in both positive and negative classifications. Thus, metrics are commonly a function of the number of true positives (TP), false positives (FP), true negatives (TN), and false negatives (FN). A confusion matrix, as in Table~\ref{tab:confusion}, reports counts or proportions for each of these.

\begin{table}[htbp]
\centering
\caption{Confusion Matrix Setup}
\label{tab:confusion}
\begin{tabular}{lcc}
\toprule
 & \multicolumn{2}{c}{Machine} \\
\cmidrule(lr){2-3}
Human & Positive & Negative \\
\midrule
Positive & TP & FN \\
Negative & FP & TN \\
\bottomrule
\end{tabular}
\par\smallskip
{\footnotesize \textit{Note.} TP = True Positive, FN = False Negative,
FP = False Positive, TN = True Negative.}
\end{table}

The relationship between the quantities in a confusion matrix is summarized using one or more of several metrics: accuracy (the proportion of correctly classified texts), recall (the proportion of positive cases identified by the model), precision (the probability that a given case is positive if identified as positive by the model), and specificity (the proportion of negative cases identified by the model), formally defined:
\[ \mathrm{Accuracy} = \frac{TP + TN}{TP + FP + FN + TN} \]
\[ \mathrm{Recall} = \frac{TP}{TP + FN} \]
\[ \mathrm{Precision} = \frac{TP}{TP + FP} \]
\[ \mathrm{Specificity} = \frac{TN}{TN + FP} \]

When presented alone, each of these has shortcomings \citep{japkowicz_why_2006}. Accuracy is inappropriate in the case of category imbalance, where a high value could be obtained despite never or rarely identifying the category of interest. Reporting recall and precision in tandem addresses this problem by providing information on the proportion of positive cases identified, and the proportion of model-identified positive cases that are indeed positive. Still, precision and recall do not provide any information on proportion of true negatives identified, an issue addressed by specificity.

As defined, each of the above assumes a true classification, likely provided by a human. However, acknowledging reasonable disagreements between even experts \citep{shaffer_how_2021}, the possibility of human error, and the impressive capabilities of LLMs \citep{gilardi_chatgpt_2023}, researchers evaluating LLM classifications may prefer to refrain from categorizing the human classification as the decisive truth when there are disagreements. (This is less the case with supervised learning where the model aims to directly replicate human classifications, thus requiring that the human classifications are treated as gold standard.) Even if analysts are agnostic on which classification is correct in the case of disagreements, measuring disagreements remains an important aspect of validation, analogous to measuring agreement between humans \citep{gwet_handbook_2014}. Analysts in this case may report percent agreement (which is subject to the same shortcomings as accuracy), or corollaries to recall and specificity: percent positive agreement (PPA) and percent negative agreement (PNA; \citealp{han_determination_2022}), defined using the disagreement matrix below (Table~\ref{tab:disagree}) as:
\[ \text{PPA} = \frac{2a}{2a + b + c} \quad \text{and} \quad
   PNA = \frac{2d}{2d + b + c}. \]

\begin{table}[htbp]
\centering
\caption{Disagreement Matrix}
\label{tab:disagree}
\begin{tabular}{lcc}
\toprule
 & \multicolumn{2}{c}{Classification 1} \\
\cmidrule(lr){2-3}
Classification 2 & Positive & Negative \\
\midrule
Positive & a & b \\
Negative & c & d \\
\bottomrule
\end{tabular}
\end{table}

\subsubsection{Categories \texorpdfstring{$k \geq 3$}{k >= 3}}

When text classification is used to categorize a text into greater than two possible categories, each category \emph{k} is associated with its own confusion/disagreement matrix. The metrics above, then, remain unchanged in definition, and can be calculated for each category. To aggregate across categories, however, we can take an average, either giving equal weight to each category (macro-averaging, e.g., generating macro-recall) or weighting by the number of texts in a category (micro-averaging, e.g., generating micro-recall). The latter is more representative of the sample (as each text counts equally) but can easily become dominated by large categories \citep{rainio_evaluation_2024}.

\subsubsection{Summary Metrics}

Comprehensive evaluation of a text classifier requires obtaining an understanding of the multiple types of error or disagreement, among positives or negatives (and, when relevant, within multiple categories). However, a single summary metric simplifies the selection between classifiers (or among hyper-parameters, text representations, prompts, etc.). Metrics like F1, the area under the Receiver Operating Characteristic curve and Cohen's Kappa ($\kappa$) serve this function well.

F1 is defined as the harmonic mean of precision and recall, formally:
\[ F1 = \frac{2 \times Precision \times Recall}{Precision + Recall}, \]
allowing an analyst to compare multiple classifiers, some with higher precision than recall and vice versa. ROC is relevant when using a classifier that outputs probabilities and provides the probability that, if given a randomly selected positive and negative example, the model will rank the positive case higher than the negative case \citep{ling_auc_2003}. Cohen's Kappa $(\kappa) = \frac{Agreement - p_{e}}{1 - p_{e}}$ given $p_{e} = \frac{(a + b)(a + c) + (c + d)(b + d)}{(a + b + c + d)^{2}}$, provides a measure of agreement, beyond that which can be attributed to random chance, $p_{e}$. Like PPA and PNA, $\kappa$ is agnostic to ``truth''; either the machine or the human could be considered the ``truth'', and the metrics would be the same.

While summary metrics are particularly well-suited to the automated empirical selection of hyper-parameters, as they provide a single metric to optimize for, they are less directly informative about the error behavior of the model when employed \citep{green_algorithmic_2020, mitchell_model_2019}. For example, reporting only an F1 score may hide the fact that a model has high precision but low recall (or vice versa), a characteristic which would impact the degree to which it is useful, and which impacts the appropriate interpretation of its classifications. 

\section{Confidence Intervals for Independent Data}

Reported performance metrics are a best estimate, given a sample, of an unknown true performance rate among the population of texts to which the classifier may be applied. Because of sampling variability, estimated performance is subject to random error and will vary from sample to sample. Confidence intervals provide a range of values within which the true performance rate is likely to fall.

Unfortunately, in many social science journals, performance metrics are, at best, inconsistently accompanied by measures of uncertainty \citep{anglin_addressing_2024}. The failure to report standard errors or confidence intervals around a metric treats model performance as if it is known, e.g., as if we are certain that the classifier will identify 85\% of instances of a construct. However, performance is always assessed on a sample of texts, with accompanying confidence interval widths depending on the number of texts, classifier performance, and category prevalence.

Unfortunately, confidence interval width also depends on the method of estimation, with some estimation approaches resulting in tighter, but inaccurate, intervals, and others resulting in wider but reliable intervals. The aim, of course, is to obtain intervals that accurately reflect uncertainty---e.g., so that a 95\% confidence interval contains the true metric in approximately 95\% of repeated samples---while remaining as narrow as possible. We define several methods of interval calculation below. 

\subsection{Analytic Intervals}

We begin with the textbook definition of a confidence interval for a binomial proportion, the Wald confidence interval, sometimes called the normal approximation \citep{raschka_model_2018}. Let $X$ denote the number of instances (``successes'' in binomial terms) for a sample size $n$ (trials; e.g., total number of texts). Then let $\widehat{p} = \frac{X}{n}$ represent the sample proportion. Using the asymptotic normality of the sample proportion, the $100(1-\alpha)\%$ confidence interval for $p$ is:
\[ \widehat{p} \pm z_{\alpha/2}\sqrt{\frac{\widehat{p}(1 - \widehat{p})}{n}} \]

Consider the following example. If there were a total of 2000 texts and 300 cases of the construct of interest, the 95\% Wald confidence interval surrounding the proportion would be $[0.134, 0.166]$. Of course, however, the aim here is not to estimate a confidence interval surrounding the proportion of positive instances of a construct, but rather the model's identification of such instances. This requires modifying the definitions of $n$ and $p$.

In the calculation of recall for the example above, $n$ becomes 300, the number of positive human classifications; $\widehat{p}$ becomes the estimated proportion of those texts which the model also identifies. Assume an estimated recall of 0.88 (the model identifies 264 of the 300 human-identified instances), the Wald confidence interval surrounding that metric would be $[0.843, 0.917]$. However, the textbook Wald estimator is known to provide poor and erratic interval estimates when $\widehat{p}$ is either high or low, particularly with small sample sizes \citep{agresti_approximate_1998}.

A commonly accepted alternative which addresses this shortcoming, is the Wilson interval, derived by inverting the score test for $p$ \citep{agresti_approximate_1998}. Let $z = z_{\alpha/2}$, then, the Wilson confidence interval for p is defined as:
\[ \frac{n\widehat{p} + \frac{1}{2}z^{2}}{n + z^{2}}
   \pm z\sqrt{\frac{n\widehat{p}(1 - \widehat{p}) + \frac{1}{4}z^{2}}
   {(n + z^{2})^{2}}} \]

Notably, the midpoint of the interval is no longer $\widehat{p}$ but can instead be understood as a weighted average of $\widehat{p}$ and $\tfrac{1}{2}$ with added proxy observations ($z$ of $1.96$, $z^{2} \approx 4$), half successes, half failures \citep{kahouadji_comprehensive_2025}, seen with some reformulation, $\widehat{p}\left(\frac{n}{n + z^{2}}\right) + \frac{1}{2}\left(\frac{z^{2}}{n + z^{2}}\right)$ \citep{agresti_approximate_1998}. The lower and upper bounds similarly incorporate proxy observations, with the number depending on the specified confidence level. In the above example, the Wilson confidence interval is both a little wider, and a little lower than the Wald interval, now $[0.838, 0.912]$. If the sample size were much lower, the difference would be larger. For example, at 30 positive observations and a recall of 0.88, the Wald interval is $[0.763, 0.996]$ while the Wilson interval is $[0.719, 0.955]$.

Two additional analytic intervals for proportions may be useful. First, Agresti and Coull create a simplified version of the Wilson interval, generated by simply adding a count of two successes and two failures to the Wald interval \citep{agresti_approximate_1998}:
\[ \widetilde{p} \pm z\sqrt{\frac{\widetilde{p}(1 - \widetilde{p})}{\widetilde{n}}} \]
This small change substantially improves the behavior of the interval at proportions near 0 and 1. Second, the Clopper-Pearson interval \citep{clopper_use_1934} is an exact interval guaranteed to cover the true p in $100(1 - \alpha)\%$ of binomial samples, albeit conservatively \citep{agresti_approximate_1998}.

\subsection{Bootstrapped Intervals}

Bootstrapping provides a non-analytic alternative to the calculation of standard errors. The approach is particularly compelling when there is no readily available formula for the standard error of a statistic. Most notably, this is the case for the F1 metric, which is not a binomial proportion (and thus, none of the above specifications of the Wald, the Wilson, Agresti-Coull, or Clopper-Pearson is applicable), but the harmonic mean of precision and recall.

When bootstrapping, we simulate statistical inference. Given an i.i.d. sample of $n$ individuals, we: 1) draw a random sample of $n$ individuals with replacement, creating $\mathbf{x}^{*} = \{x_{1}^{*}, x_{2}^{*}, x_{3}^{*}, \ldots, x_{n}^{*}\}$; 2) calculate the metric of interest $\widehat{m}^{*}$ (e.g., $\widehat{F1}^{*}$); 3) Repeat this process a large number of times, say $B = 1000$, generating 1000 bootstrap replicates of $\widehat{m}^{*}$ \citep{tibshirani_introduction_1993}. With this basic setup, several options for confidence intervals exist.

\paragraph{Percentile Interval.} One of the most straightforward bootstrapped confidence intervals is simply the range in the middle $1 - \alpha$ of a bootstrap distribution; e.g., for a 95\% confidence interval with 1000 bootstraps, the 25\textsuperscript{th} and 975\textsuperscript{th} largest of the bootstrap replicates serve as an estimate of the 95\% confidence interval \citep{efron_nonparametric_1981}. Unfortunately, this approach is often inaccurate and erratic at small to moderate sample sizes when proportions near 0 or 1 \citep{pires_interval_2008}.

\paragraph{BCa Interval.} The bias-corrected and accelerated interval (BCa; \citealp{tibshirani_introduction_1993}) uses percentiles adjusted by a bias parameter and an acceleration parameter. The bias parameter depends on the proportion of bootstrap statistics falling below the observed statistic observed from the original sample while the acceleration parameter depends on the skewness of empirical influence function, estimated using jackknife leave-one-out estimates \citep{hesterberg_bootstrap_2011, tibshirani_introduction_1993}.

\paragraph{Pseudo-Count Regularized Interval.} Finally, we propose a simple adjustment to the bootstrap to address the erratic nature of bootstrapped intervals when the sample size is small and $p$ is either high or low. The intuition and implementation of the interval is very similar to the analytic Agresti-Coull interval: the original sample is augmented with an equal number of pseudo-successes and pseudo-failures. Then, the bootstrapped percentile interval is calculated as usual, with some bootstrapped samples including one or more of the pseudo-observations. More formally, we augment the i.i.d. sample $\mathbf{x}$ of $n$ observations, $\mathbf{x} = \{x_{1}, x_{2}, x_{3}, \ldots, x_{n}\}$ with $\lambda$ pseudo-successes and pseudo-failures, generating:
\[ \mathbf{x}_{\text{aug}} = \Big\{ x_{1}, \ldots, x_{n},
   \underbrace{1, \ldots, 1}_{\lambda \text{ pseudo-successes}},
   \underbrace{0, \ldots, 0}_{\lambda \text{ pseudo-failures}} \Big\}. \]
E.g., if $\lambda = 1$, then $\mathbf{x}_{\text{aug}} = \{x_{1}, x_{2}, x_{3}, \ldots, x_{n}, 1, 0\}$. Then we: 1) draw a random sample of $n + 2\lambda$ observations with replacement from $\mathbf{x}_{\text{aug}}$, creating $\mathbf{x}_{\text{aug}}^{*} = \{x_{1}^{*}, x_{2}^{*}, x_{3}^{*}, \ldots, x_{n + 2\lambda}^{*}\}$; 2)  calculate the metric of interest $\widehat{m}^{*}$ (e.g., $\widehat{F1}^{*}$); 3) Repeat this process $\beta$ times, generating $\beta$ bootstrap replicates of $\widehat{m}^{*}$. 4) Identify the appropriate percentiles to generate the interval.

Like in the Wilson and Agresti-Coull intervals, the pseudo-count regularized bootstrap interval for a proportion is centered around $\widetilde{p}$ weighted towards 0.5 and the magnitude of the weight depends on $n$. When $n$ is large, $\lambda$ will be comparatively small and the bootstrap samples will be dominated by true observations. On the other hand, when $n$ is small, the center will be pulled more strongly towards 0.5. When applied to performance metrics, this reduces the lower bound of an interval with a high estimated performance. To generate an interval around $F1$, which is not a single proportion but the harmonic mean of two, we augment the sample with $\lambda$ pseudo-observations in each quadrant of the confusion matrix before applying the bootstrap procedure. Thus, both precision and recall are regularized as above.

\section{Coverage Under Independence}

This section demonstrates the coverage of 95\% confidence intervals resulting from the Wald, Wilson, Agresti-Coull, and Clopper-Pearson confidence intervals (the analytic methods), as well as percentile, BCa, and pseudo-count regularized bootstraps. We define coverage as the proportion of samples---drawn from a Bernoulli distribution with a true population proportion $p$---where the 95\% confidence interval contains $p$, i.e., $\Pr\{p \in CI\}$. In other words, take a population recall of 0.8 (i.e., where, if we could assess our classifier on the full population of human-coded positive texts, a proportion of 0.8 of these positive cases would be identified by the model). Then, we want to know, for each of the interval methods assessed, what proportion of random samples of size $n$, would result in a 95\% confidence interval which contains 0.8. A 95\% confidence interval method should reach approximately 0.95 coverage.

We test coverage varying n from 10 to 250 and varying $p$ to include 0.5, 0.6, 0.7, 0.8, 0.9, and 0.95. When calculating bootstrapped confidence intervals, we generate 2000 bootstrapped samples. If any bootstrapped sample does not contain a positive case, that sample is excluded from the calculation of the confidence intervals. For the pseudo-count bootstrap, we assess coverage with $\lambda = 1$ and $2$ (the number of successes and failures added; e.g., $\lambda = 1$, indicates one pseudo-success and one pseudo-failure). This coverage study replicates the work of \citet{agresti_approximate_1998}, \citet{pires_interval_2008}, \citet{kahouadji_comprehensive_2025}, among others, in its comparison of confidence interval coverage, but with a focus on values of $p$ likely obtained with performance metrics. It is more novel in its inclusion of a simple and intuitive approach to addressing low coverage for bootstrapped intervals calculated on small samples with high proportions (via pseudo counts).

\subsection{Coverage for Proportion-Based Metrics}

\begin{figure}[htbp]
\centering
\includegraphics[width=\linewidth]{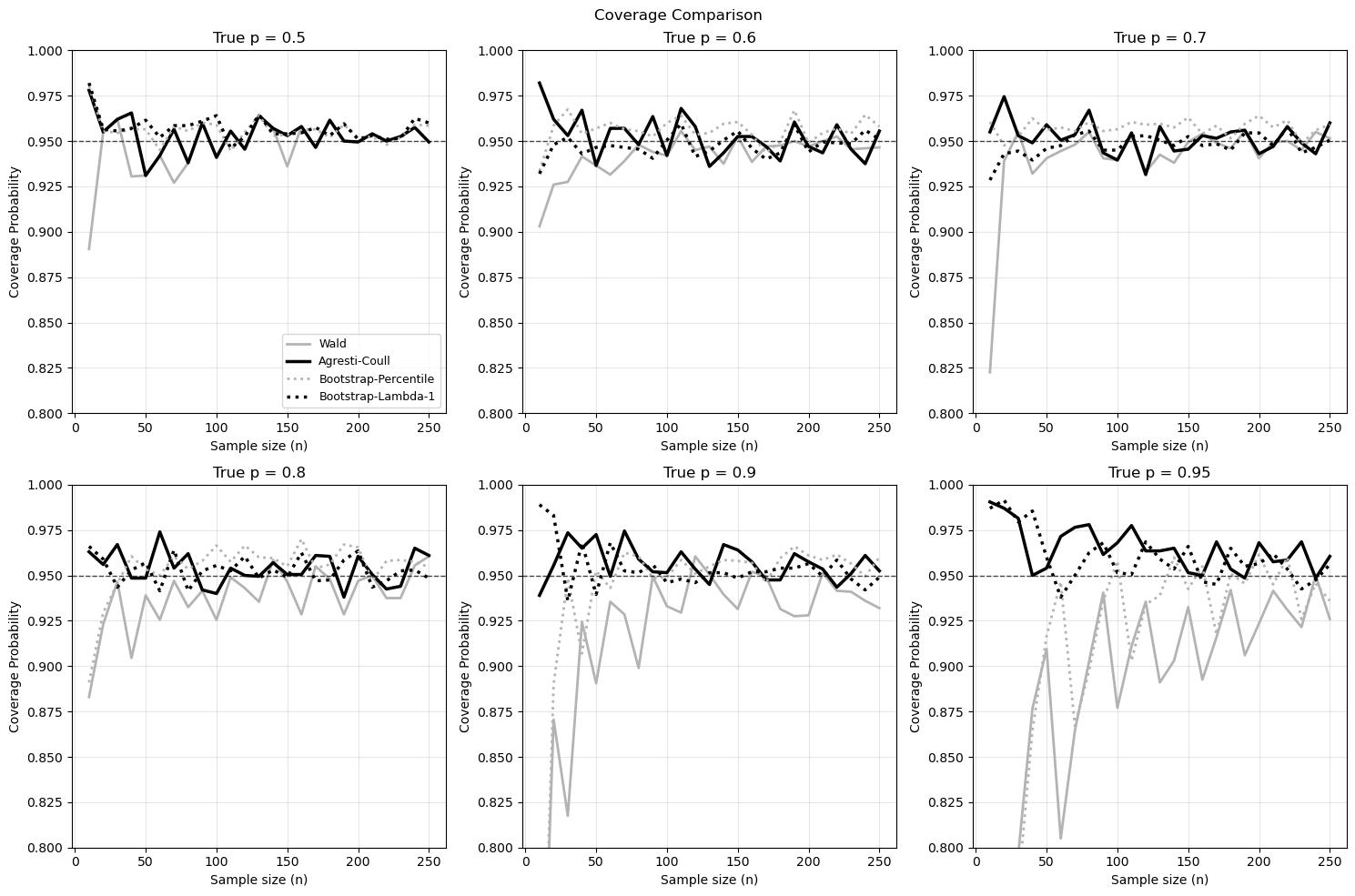}
\caption{Coverage Rates for Interval Calculation Methods}
\label{fig:fig1}
\end{figure}

A comprehensive set of our coverage results for analytic intervals may be found in Appendix~\ref{app:A}, and bootstrapped intervals in Appendix~\ref{app:B}. Our key conclusions are summarized in Figure~\ref{fig:fig1} which show coverage results for the two interval methods that demonstrate the worst behavior (despite being the most commonly employed in practice; the Wald intervals and the basic percentile bootstrap) alongside two strong alternatives (the Agresti-Coull interval and a pseudo-count regularized bootstrap with $\lambda = 1$).

As expected, the Wald interval demonstrates undesirable properties. At higher proportions and sample sizes below 50, coverage is often below 0.9 with some coverage rates reaching 0.6 and below. Further, coverage is erratic with some sample sizes producing dramatic drops in coverage (see Figure~\ref{fig:fig1}). Even at proportions of 0.5, 0.6, and 0.7, when sample sizes are below 50 coverage is often lower than the nominal level; around 0.925 for a 95\% confidence interval. On the other hand, the Clopper-Pearson, Wilson, and Agresti-Coull intervals all offer substantial improvements. For example, Agresti-Coull intervals consistently hover around or above 0.95 coverage, even at small sample sizes and extreme values of $p$. Coverage rates for Clopper-Pearson and Wilson intervals are similar (see Appendix~\ref{app:A}), though, rates for Clopper-Pearson are always at least 0.95 (as expected). Overall, each of these three intervals (Agresti-Coull, Wilson, or Clopper-Pearson) is a reasonable option. The Wilson and Agresti-Coull intervals have a slight advantage in terms of interval width (see Figure~\ref{fig:figA5} for a comparison of interval half-widths) and a slight disadvantage in terms of coverage.

Similar to the Wald interval, the basic percentile bootstrap also demonstrates insufficient and erratic coverage, particularly at high proportions. When $p = 0.95$, confidence interval coverage rarely reaches 0.95. Further, like the Wald interval, there are values of $n$ that result in dramatic drops in coverage, and coverage is more often slightly below 0.95 than above. The BCa addresses this last problem, pushing coverage higher but problems remain when the sample size is low while the proportion is high (see Appendix~\ref{app:B}). The pseudo-count regularized bootstrap interval with $\lambda = 1$ (augmenting one success and one failure to the sample before bootstrapping; see Appendix~\ref{app:B} for an exploration of the effectiveness of the regularizing with $\lambda = 1$ versus 2) addresses all three problems. Coverage is consistently at or above 0.95 for all presented proportions and sample sizes and interval width is no wider than the BCa intervals (see Appendix~\ref{app:B}).

Overall, the Wilson, Agresti-Coull, Clopper-Pearson, and pseudo-count regularized interval all provide appropriate coverage. We prefer the Agresti-Coull for its simplicity and limited computational cost (compared to bootstrapping). However, the pseudo-count bootstrap offers an alternative for metrics like F1 where no analytic interval is readily available. We explore coverage in this scenario below.

\subsection{Coverage for F1}

We evaluate the empirical coverage of bootstrap confidence intervals for F1 under population conditions that vary true precision and true recall at values of 0.50, 0.60, 0.70, 0.80, 0.90, and 0.95. Prevalence is set to 0.10 (indicating 10\% of the population observations are in the positive class), allowing us to assess coverage in the case of imbalanced classes. For each condition, we convert target precision and recall values into population probabilities in each quadrant of the confusion matrix. Then, the true F1 score for each population is calculated from the population precision and recall. For each population condition, coverage is assessed across sample sizes ranging from 30 to 500, increasing by one from 30 to 50, and then by 10 from 60 to 500. For each sample size and population condition, 1,000 independent samples were generated, each with an associated confusion matrix with counts of true positives, false positives, true negatives, and false negatives.

For each simulated sample, we construct a pseudo-count regularized bootstrapped interval with $\lambda = 1$ by adding one count to each quadrant of the matrix. Then, to generate a bootstrapped sample, $n + 4\lambda$ observations (categorized as TP, FP, TN, or FN) are randomly selected with replacement. Precision, recall, and F1 are calculated for each of 2000 bootstrapped samples and we take the 2.5th and 97.5th percentiles to generate the 95\% confidence interval. Coverage is estimated as the proportion of simulated sample intervals containing the known population F1 score.

\paragraph{F1 Coverage Results.}

\begin{figure}[htbp]
\centering
\includegraphics[width=0.62\linewidth]{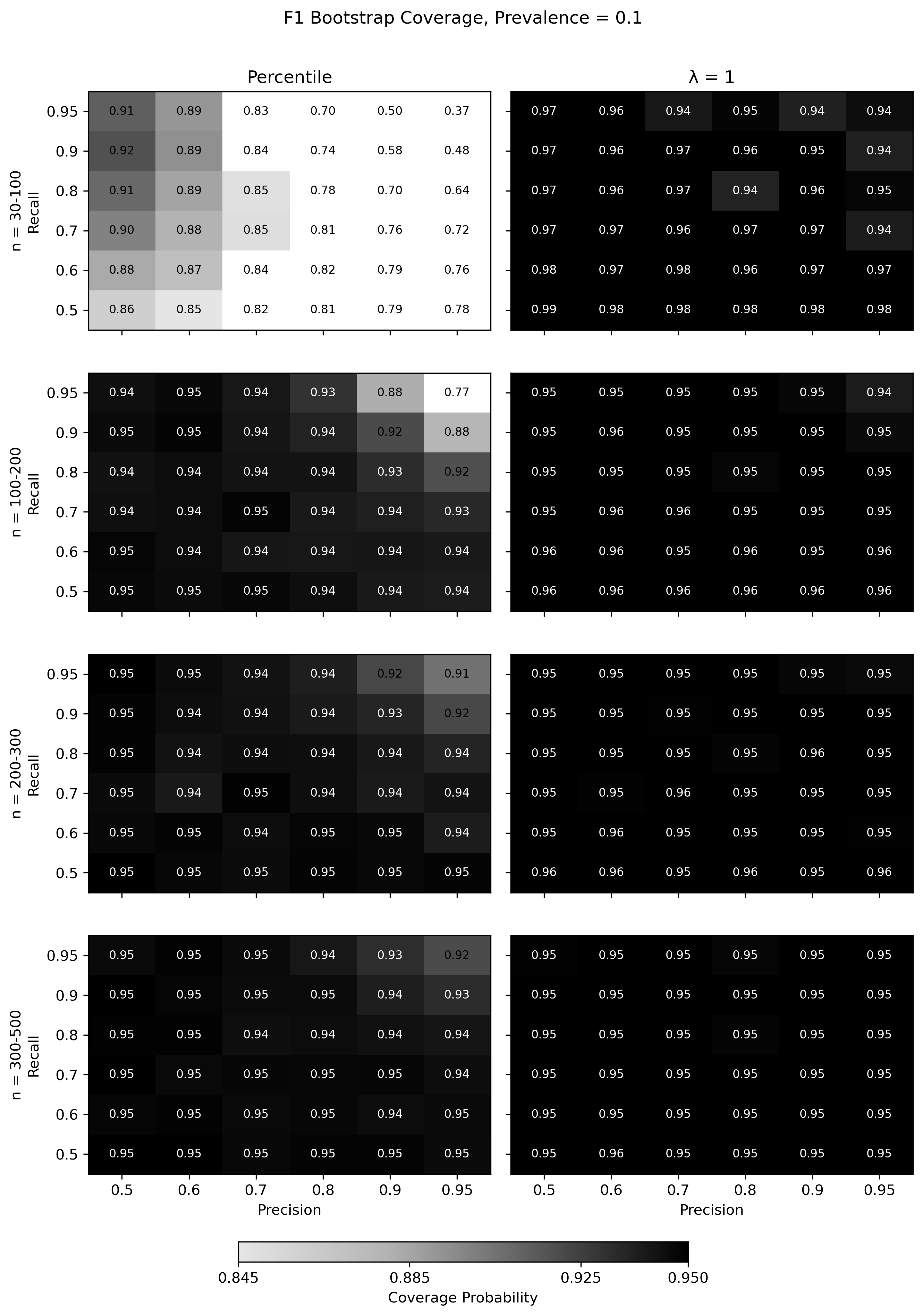}
\caption{Coverage Rates for the Percentile Bootstrap and Pseudo-Count
Regularized Bootstrap ($\bm{\lambda = 1}$) at a Prevalence of 0.10}
\label{fig:fig2}
\end{figure}

Figure~\ref{fig:fig2} displays the empirical coverage of 95\% bootstrap confidence intervals for F1 across the simulated populations where construct prevalence = 0.10. Each row corresponds to a sample size range and each heatmap cell corresponds to one combination of true precision and true recall, averaged across the relevant sample sizes. The left-hand panel shows coverage with the percentile bootstrap and the right-hand shows coverage with the pseudo-count regularized bootstrap where $\lambda = 1$. Shading represents the estimated coverage probability, with darker values indicating coverage closer to or above the desired 0.95 level. Higher coverage is observed with the pseudo-regularized bootstrap than the percentile bootstrap. For example, when $n = 30\text{--}100$ the non-regularized bootstrap rarely reaches even 0.90 coverage. Note here that 30 is a very small sample size, given we simulate 10\% construct prevalence. However, even when sample sizes are between 100--300, coverage is often still below 0.95. On the other hand, coverage for the regularized bootstrap is almost always at the desired 0.95 level and is consistently better than that of basic percentile bootstrap.

\subsection{Sample Size Considerations}

\begin{figure}[htbp]
\centering
\includegraphics[width=\linewidth]{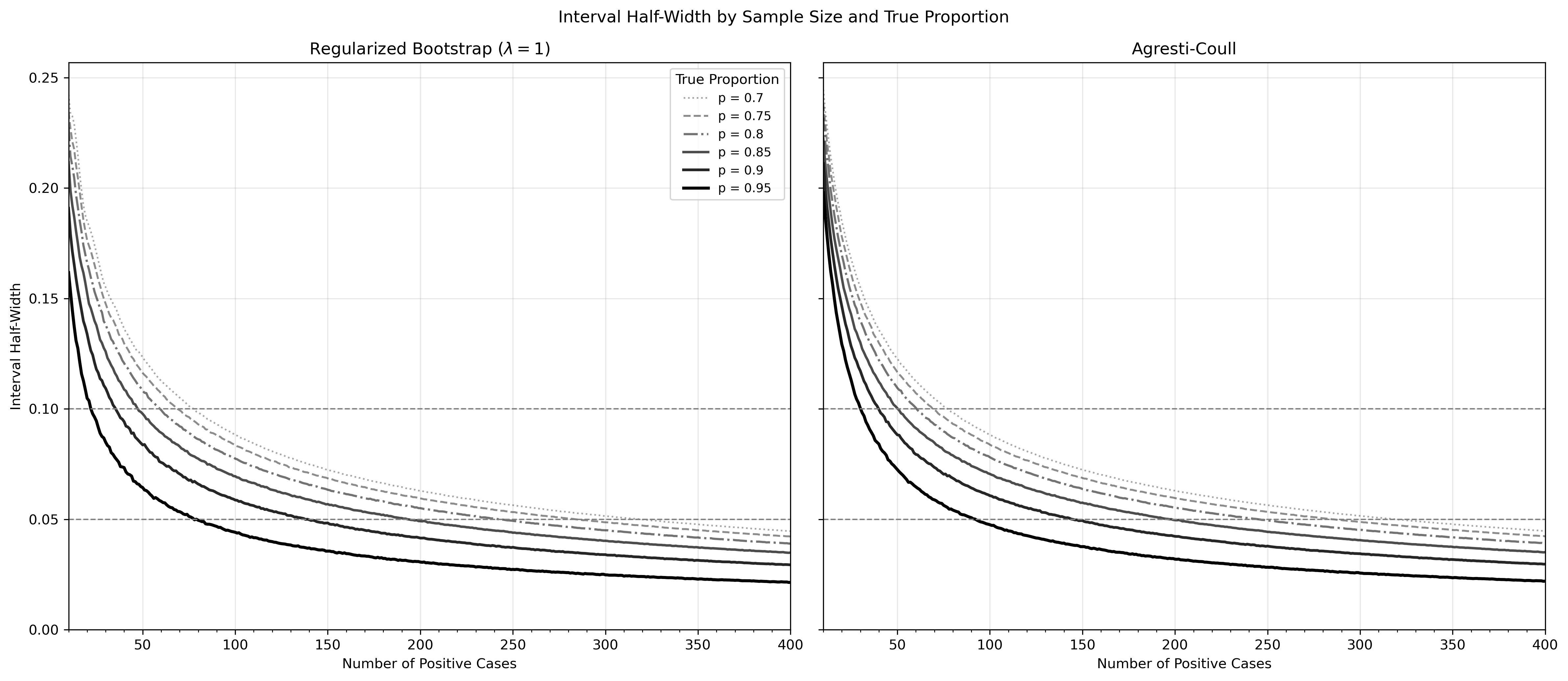}
\caption{Relationship between Confidence Interval Half-Width and Sample Size}
\label{fig:fig3}
\end{figure}

Figure~\ref{fig:fig3} displays the relationship between the number of cases used to estimate performance and the half width of a 95\% confidence interval, with varying performance rates represented. The figure reinforces two key behaviors relevant to sample size planning. First, higher performance rates result in tighter confidence intervals, and this relationship grows stronger as performance gets higher (e.g., the difference in half-widths between $p = 0.90$ and $p = 0.95$ is greater than the difference in half-widths between $p = 0.70$ and $p = 0.75$). Therefore, the needed validation sample size depends heavily on the expected performance of the model. Second, interval half-width decreases with sample size in an inverse square relationship. This relationship means it is much more costly to move from a half-width of 0.05 to 0.025 than from 0.20 to 0.15.

\begin{table}[htbp]
\centering
\footnotesize
\caption{Necessary Sample Size for Desired Interval Half-Width, and True
Proportion-Based Performance}
\label{tab:samplesize}
\begin{tabular}{ll cccc cccc}
\toprule
 & & \multicolumn{4}{c}{Regularized Bootstrapped} & \multicolumn{4}{c}{Agresti-Coull} \\
\cmidrule(lr){3-6}\cmidrule(lr){7-10}
CI & Proportion & $\le 0.025$ & $\le 0.05$ & $\le 0.075$ & $\le 0.1$
              & $\le 0.025$ & $\le 0.05$ & $\le 0.075$ & $\le 0.1$ \\
\midrule
\multirow{6}{*}{90\%}
 & 0.70 & 904 & 224 & 99 & 55 & 907 & 226 & 99 & 55 \\
 & 0.75 & 807 & 200 & 88 & 48 & 809 & 201 & 89 & 49 \\
 & 0.80 & 689 & 170 & 75 & 42 & 691 & 172 & 76 & 43 \\
 & 0.85 & 549 & 137 & 61 & 34 & 552 & 138 & 62 & 35 \\
 & 0.90 & 389 & 99  & 44 & 25 & 392 & 100 & 46 & 27 \\
 & 0.95 & 212 & 56  & 28 & 18 & 216 & 61  & 31 & 20 \\
\midrule
\multirow{6}{*}{95\%}
 & 0.70 & 1282 & 320 & 140 & 78 & 1288 & 320 & 140 & 78 \\
 & 0.75 & 1145 & 284 & 125 & 69 & 1149 & 285 & 126 & 70 \\
 & 0.80 & 978  & 243 & 107 & 60 & 983  & 245 & 108 & 61 \\
 & 0.85 & 778  & 194 & 85  & 48 & 786  & 199 & 89  & 51 \\
 & 0.90 & 547  & 138 & 62  & 36 & 562  & 145 & 69  & 40 \\
 & 0.95 & 297  & 78  & 38  & 23 & 314  & 93  & 48  & 31 \\
\bottomrule
\end{tabular}
\end{table}

Table~\ref{tab:samplesize} provides similar information in table form, alongside a 90\% interval, for ease of reference. Note that here, and in the figure above, in the context of validating binary text classifiers, the analytic sample size is the number of observations used to estimate the metric, not the total number of labelled observations. For recall, this would be the number of positive human classifications, for precision, the number of positive model classifications. In the table, if we want to be able to estimate a regularized bootstrapped 90\% confidence interval no wider than 0.05 on either side and aim to generate a classifier capable of identifying 90\% of positive cases (i.e., recall 0.90), we would need 99 positive cases. If only a quarter of the texts are labelled by humans as positive, then the validation data would need to be 396 texts for a confidence interval half-width of 0.05. Again, analytic sample size requirements increase if we expect a classifier to reach a lower level of performance. At an assumed performance of 0.80, for example, 170 positive cases would be needed.

\section{Confidence Intervals for Nested Data}

Studies of text data are often hierarchical in nature, with multiple texts collected from the same individual. For example, we may have utterances (which may or may not reflect our construct of interest), nested within therapists. Such texts are not independent. For example, utterances produced by the same therapist likely share many characteristics as the therapist has their own style of speech used when interacting with clients. Such nested datasets produce two potential limitations in text classification. First, the confidence interval estimation procedures described above assume independent binomial trials \citep{dean_evaluating_2015, diciccio_review_1988}. This assumption can result in an underestimation of uncertainty, resulting in confidence intervals that are inappropriately narrow \citep{saravanan_application_2020}. Second, if data are split into training and testing sets at the text level, the model may benefit from learning the ``ungeneralizable idiosyncrasies'' of individuals \citep[p.~13]{anglin_addressing_2024}. Thus, performance is likely to be lower when applied to new, unseen individuals. In such a scenario, it is considered best practice to conduct a train:test split at the highest level of the hierarchical data, such that no single speaker's utterances are in both the training and testing data \citep{anglin_addressing_2024}.

On the other hand, if the model will never be applied to unseen individuals, such a split may not be necessary. The key question is: Does the available corpus represent a sample of texts to which the classifier may be applied, or does it represent the full population? In other words, will the classifier be applied to texts outside those from which the labelled data were sampled? If not, it is not necessarily overfit for a classifier to learn person-specific idiosyncrasies within the dataset; these idiosyncrasies are relevant. On the other hand, if a classifier will be applied beyond the current available sample, the analyst will want to know that it can generalize to entirely unseen individuals. In this case, splitting should occur at the individual level, with confidence intervals that reflect this sampling scheme. 

\subsection{The Design Effect}

When texts are produced by the same individual, there is reason to believe that there exists some homogeneity of performance within that individual: i.e., the model may be better suited at identifying the concept within one individual's way of writing or speaking compared to another. To the extent that there is homogeneity in performance, this violates the assumption of independent and identically distributed (i.i.d.) observations. The degree to which this matters in estimates of variance is known as a design effect, first proposed to understand survey variance under cluster designs \citep{kish_survey_1965}: $DEFF = \frac{Var_{\text{actual}}}{Var_{\text{srs}}}$. The numerator is the variance under the actual, hierarchical design (e.g., in the original framework, individuals sampled within neighborhoods for surveys; here, texts sampled within individuals for text classification) and the denominator is the variance under simple random sampling \citep{kish_survey_1965}. A design effect greater than one indicates the increase in variance (and the loss of statistical precision) due to clustering. Expressed in terms of standard errors relevant for confidence interval width: $s.e._{\text{act}} = s.e._{\text{srs}} \times \sqrt{\text{DEFF}}$ such that the confidence interval width is multiplied by $\sqrt{\text{DEFF}}$. With a 95\% confidence interval and a design effect of 1.5, $\sqrt{1.5} \approx 1.225$, for example, the confidence interval will be approximately 22.5\% wider than under simple random sampling.

The magnitude of the design effect is dependent on the degree of homogeneity within clusters (or here, the homogeneity of performance within texts produced by the same individual), measured via the intra-class correlation, $\rho = \frac{Var_{\text{between}}}{Var_{\text{between}} + Var_{\text{within}}}$, the proportion of total variance due to variance between clusters. (See \citealp{wu_comparison_2012} for a comparison of several approaches to estimating ICC.) In the case of text classification, a higher proportion would indicate a higher degree of within-individual performance homogeneity. If cluster sizes are equal, the design effect can be approximated by $1 + \rho(B - 1)$, i.e., the ICC or $\rho$, multiplied by the number of observations within a cluster, $B$, less one \citep[p.~162]{kish_survey_1965}. If cluster sizes are unequal, substituting the average cluster size for $B$ also provides a reasonable approximation \citep{kish_survey_1965}.

\subsection{Adjusted Analytic Intervals}

When data are not i.i.d., analytic confidence intervals may be adjusted by using effective sample size, $n_{\text{eff}}$, instead of n. Effective sample size, $n_{\text{eff}}$ may be calculated \citep{korn_confidence_1998}:
\[ n_{\text{eff}} = \frac{\widehat{p}(1 - \widehat{p})}{\widehat{\text{var}}(\widehat{p})}, \]
where $\widehat{\text{var}}(\widehat{p})$ may be estimated as $\widehat{\text{var}}(\widehat{p}) = \left\{\frac{\widehat{p}(1 - \widehat{p})}{n}\right\} \times \left\{1 + \rho(B - 1)\right\}$.

Effective sample size can also be adjusted for the so called design degrees of freedom, $df_{\text{design}} = m - 1$, understood as the number of primary sampling units (here, individuals) minus the number of strata (here, 1 as the participants are not stratified) \citep{korn_confidence_1998} such that the design degrees of freedom adjusted sample size is:
\[ n_{\text{eff}}^{*} = \frac{\widehat{p}(1 - \widehat{p})}{\widehat{\text{var}}(\widehat{p})}
   \times \left\{\frac{z\left(1 - \frac{\alpha}{2}\right)}
   {t_{df_{\text{design}}}\left(1 - \frac{\alpha}{2}\right)}\right\}^{2}, \]
where $t_{df_{\text{design}}}\left(1 - \frac{\alpha}{2}\right)$ indicates the appropriate quantile with $df_{\text{design}}$ degrees of freedom \citep{dean_evaluating_2015}. (Note that Korn and Graubard, 1998, introduce a degrees of freedom adjusted sample using $t$ quantiles in both the numerator and denominator. We follow the suggestion of \citet{dean_evaluating_2015} in substituting the $z$ quantile in the numerator).

Given these values, analytic confidence intervals may be calculated using the effective sample size, $n_{\text{eff}}$, or degrees of freedom adjusted effective sample size, $n_{\text{eff}}^{*}$, in lieu of $n$.

For example, a degrees of freedom adjusted Wilson interval would be calculated:
\[ \frac{n_{\text{eff}}^{*}\widehat{p} + \frac{1}{2}z^{2}}{n_{\text{eff}}^{*} + z^{2}}
   \pm z\sqrt{\frac{n_{\text{eff}}^{*}\widehat{p}(1 - \widehat{p}) + \frac{1}{4}z^{2}}
   {(n_{\text{eff}}^{*} + z^{2})^{2}}}. \]

\subsection{Clustered and Hierarchical Bootstrap}

Bootstrapping may be designed to reflect the nested nature of data, where individuals produce multiple texts. Two common approaches exist: the hierarchical bootstrap and the clustered bootstrap. Under the cluster bootstrap, individuals are sampled with replacement and all of the (re-)sampled individuals' texts are retained in the bootstrap sample \citep{field_bootstrapping_2007, mccullagh_resampling_2000}. Under the hierarchical bootstrap, first, individuals are sampled with replacement. Then, for each individual, the analyst samples with replacement from texts produced by that individual, generating the bootstrap sample \citep{field_bootstrapping_2007, saravanan_application_2020}. In either cluster bootstrapping or hierarchical bootstrapping, the metric of interest is calculated within each bootstrap sample and the process is repeated up to the preferred number of bootstraps. From there, confidence interval calculation continues in the same manner as for the simple random sample bootstrap. In the case of the percentile interval, this means simply identifying the range in the middle $1 - \alpha$ of a bootstrap distribution. Compared to the clustered bootstrap, the hierarchical bootstrap is more conservative \citep{saravanan_application_2020}, reflecting random variation in both the sample of individuals and in the texts that they produce. In the simulations below, we test both versions.

\paragraph{Pseudo-Count Regularized Clustered/Hierarchical Bootstrap
Interval.} In addition, we also propose and test a straightforward, pseudo-count regularized version of the clustered/hierarchical bootstrap interval. As with the simple random sample version described and tested above, this is designed to address the erratic nature of bootstrapped intervals when sample sizes are small and $p$ is either high or low. Now, instead of augmenting the sample with two independent pseudo-observations, one success and one failure, we augment the sample with a pseudo-\emph{individual} comprising two pseudo-observations, one success and one failure. Or, in the case of an interval surrounding F1, we augment the sample with a pseudo-individual comprising one TP, TN, FP, and FN. Then, the hierarchical or clustered bootstrapped intervals are calculated as usual, with some bootstrapped samples including the pseudo-individual one or more times.

\section{Coverage Under Nesting}

\subsection{Simulation Design}

We assess coverage for the nested interval approaches above in a setting where the researcher splits their data at the participant level, estimating performance metrics on texts produced by a random sample of participants (representing those randomly assigned to the testing dataset). We simulate a finite population of 2000 participants with 60 texts each, for a total of 120,000 text-level observations. For each individual, $i$, we first generate a latent person-level probability $p_{i}$ representing that participant's probability of a correct classification on any text. This number is conceptualized as person-level accuracy, precision or recall. Metric probabilities are drawn from a beta distribution, $p_{i} \sim Beta(\alpha, \beta)$, parameterized to have a target mean $\mu_{\text{pop}} = 0.9$ and between-person standard deviation $\sigma_{\text{pop}} = 0.1$, resulting in an approximate ICC of 0.1, simulating a high-performance classifier with moderate ICC. Conditional on $p_{i}$, text-level outcomes were generated from $Y_{ij} \sim \text{Bernoulli}(p_{i})$, for texts $j = 1, \ldots, 60$.

We simulate two sampling schemes, one small-corpus condition where there is an average of three texts per individual (target SD = 1, minimum = 1) and one moderate-corpus with an average of 30 texts (target SD = 5). We allow the number of sampled texts to vary in order to align with common data structures when estimating precision and recall; even if there are an equal number of texts produced by each participant, their numbers of human positive classifications and/or machine positive classifications will likely vary, thus varying the number of texts they contribute to recall and precision estimation. From the population, we draw samples, varying the number of participants from 5 to 50 with 1000 samples of each population size.

The study compares unadjusted, ICC-adjusted, and bootstrap confidence intervals. The unadjusted analytic intervals treat all sampled texts as independent and include Wald, Agresti-Coull, and Wilson intervals. The ICC adjusted intervals estimate ICC using ANOVA and include two forms of adjustment---adjusting for $n_{\text{eff}}$ in one form and $n_{\text{eff}}$ \& $\text{df}_{\text{design}}$ in another, both applied to Wald, Agresti-Coull, and Wilson intervals. The bootstrap intervals include an ordinary simple random sample percentile bootstrap, a clustered bootstrap, and a hierarchical bootstrap, with pseudo-observation regularized versions of each, again setting $\lambda = 1$ (one pseudo-success; one pseudo-failure) either at the text level (for the simple random sample bootstrap) or as a pseudo-participant (for cluster and hierarchical bootstraps). Bootstrap intervals are based on 1000 bootstrap resamples.

\subsection{Coverage Results}

\begin{figure}[htbp]
\centering
\includegraphics[width=\linewidth]{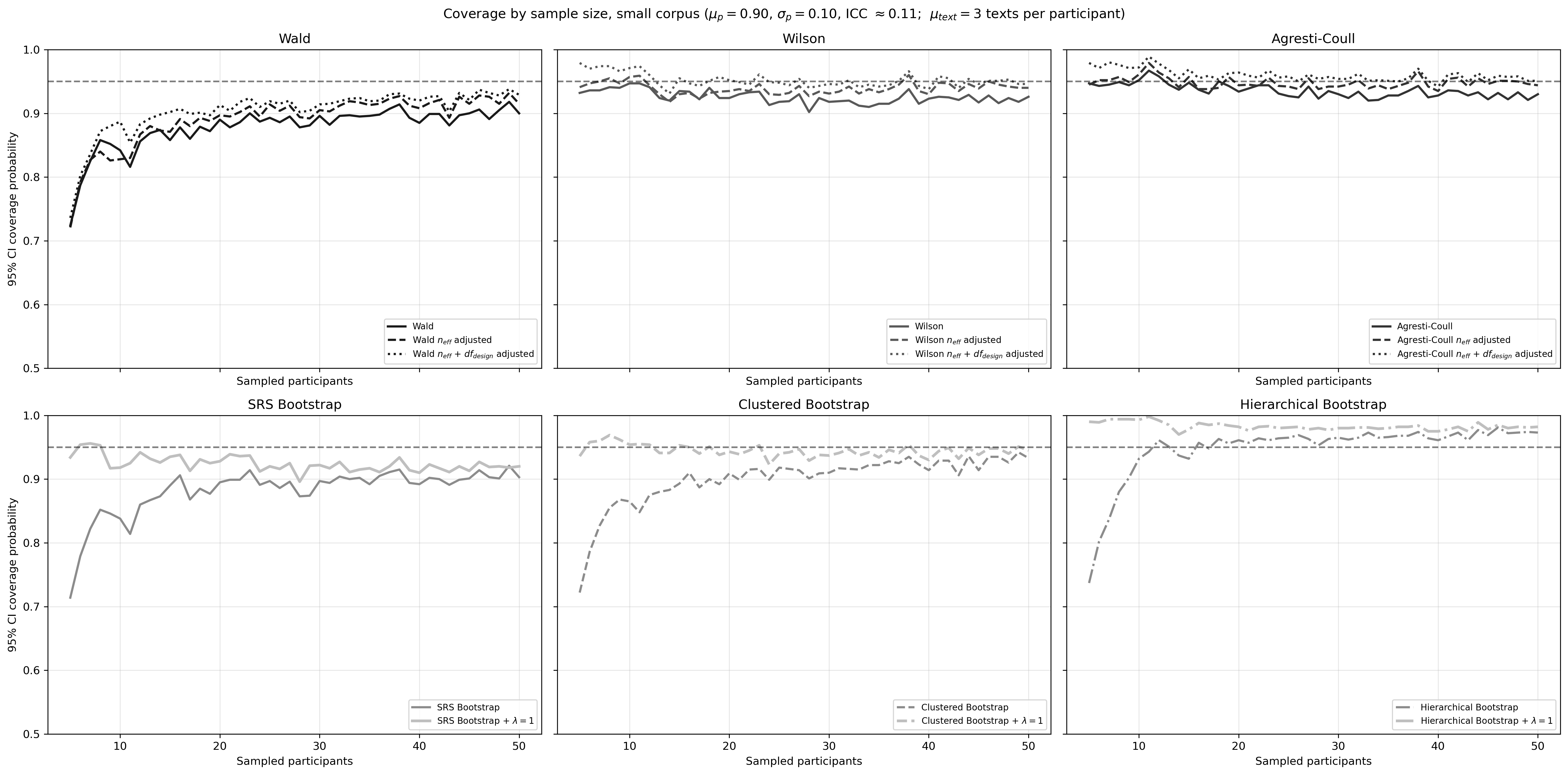}
\caption{Small Corpus Coverage ($\sim$3 Texts Per Individual)}
\label{fig:fig4}
\end{figure}

Figure~\ref{fig:fig4} plots the coverage under the $\sim$3 texts per person sampling scheme. As expected, the Wald interval displays insufficient coverage, even with ICC adjustments. However, for both Wilson intervals and Agresti-Coull intervals, ICC adjustment accounting for both $n_{\text{eff}}$ \& $\text{df}_{\text{design}}$ produces coverage around 0.95 across the sample sizes. (The difference between adjustment for $n_{\text{eff}}$ and $n_{\text{eff}}$ \& $\text{df}_{\text{design}}$ is minimal for the Agresti-Coull intervals but more pronounced at small sample sizes for the Wilson interval.) Across the bootstrap interval methods, pseudo-count regularization increases coverage particularly at very small sample sizes. Importantly, the simple random sample bootstrap consistently shows under-coverage, averaging approximately 0.92 across the sample sizes, even after regularization. On the other hand, both the regularized cluster bootstrap and the regularized hierarchical bootstrap fluctuate around 0.95, with the regularized hierarchical bootstrap consistently above.

\begin{figure}[htbp]
\centering
\includegraphics[width=\linewidth]{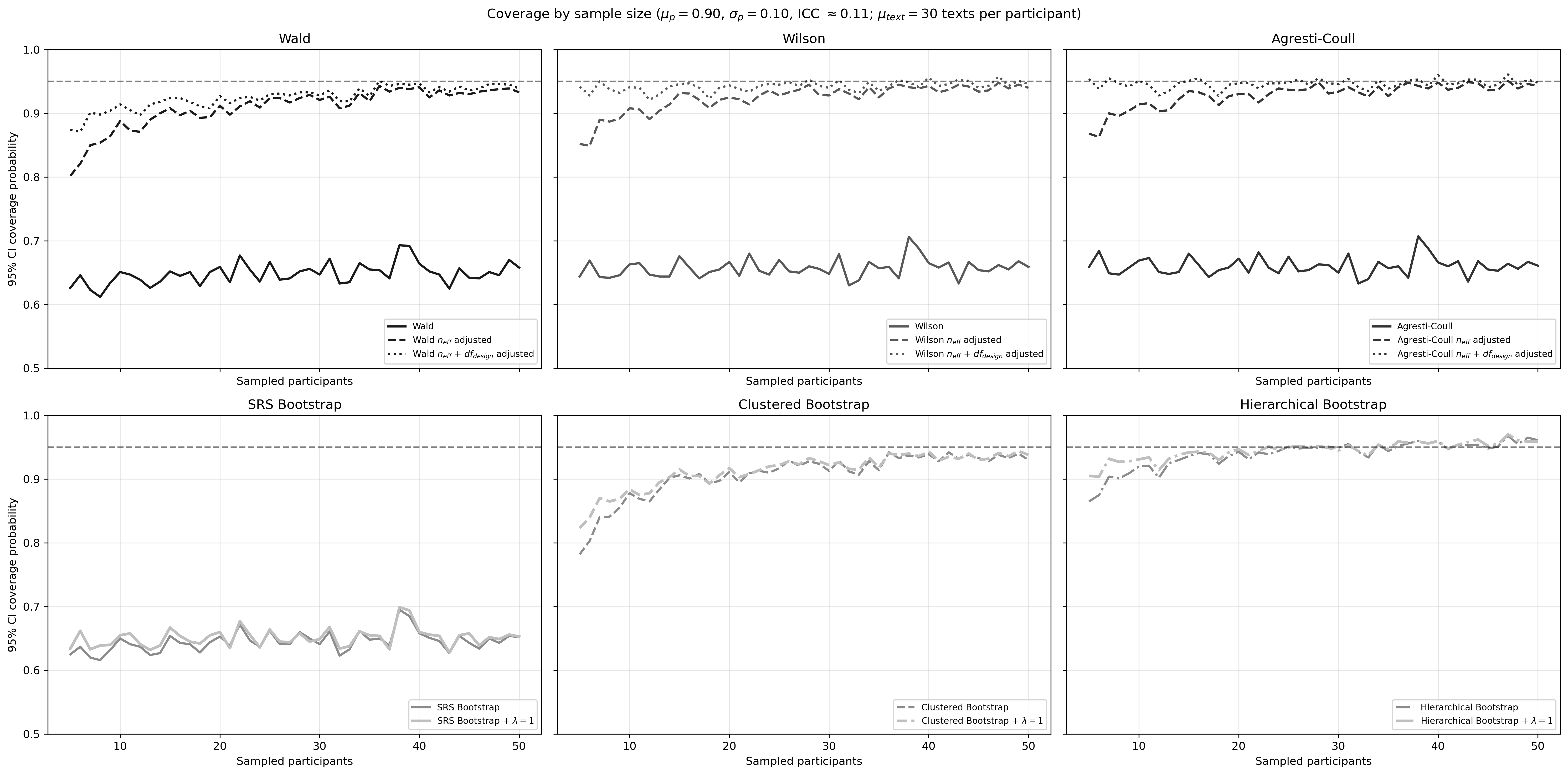}
\caption{Moderate Corpus Coverage ($\sim$30 Texts Per Individual)}
\label{fig:fig5}
\end{figure}

Results are similar but more pronounced in Figure~\ref{fig:fig5}, which displays coverage for the $\sim$30-text sampling scheme. Here, the difference between $n_{\text{eff}}$ and $n_{\text{eff}}$ \& $\text{df}_{\text{design}}$ adjustment is somewhat larger, again in favor of $n_{\text{eff}}$ \& $\text{df}_{\text{design}}$. For example, at 20 participants, $n_{\text{eff}}$ adjusted Agresti-Coull interval reaches 0.930 while the $n_{\text{eff}}$ \& $\text{df}_{\text{design}}$ reaches 0.947. The difference in coverage between the cluster and hierarchical bootstrap is more pronounced as well. Again at 20 participants, the pseudo-count regularized clustered bootstrap has an estimated coverage of 0.917 while the pseudo-count regularized hierarchical bootstrap reaches 0.947. Across all conditions, the pseudo-count regularized hierarchical bootstrap and the $n_{\text{eff}}$ \& $\text{df}_{\text{design}}$ adjusted Agresti-Coull and Wilson intervals provide approximately accurate coverage.

\begin{figure}[htbp]
\centering
\includegraphics[width=\linewidth]{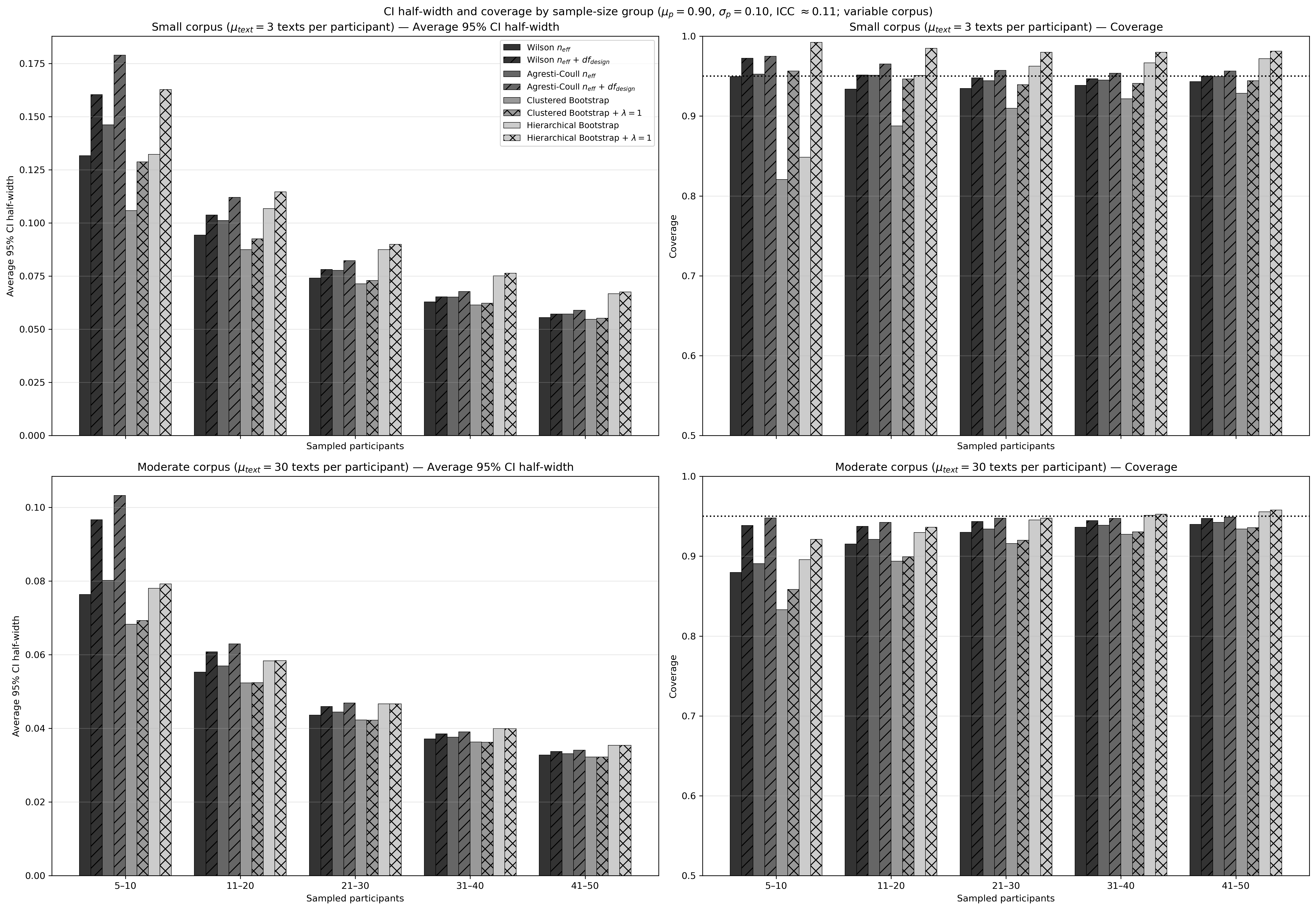}
\caption{Half-Width and Coverage Comparison for the Small-Corpus Design
($\sim$3 Texts Per Person)}
\label{fig:fig6}
\end{figure}

Figure~\ref{fig:fig6} places interval half-widths next to coverage, allowing for the assessment of trade-offs between interval precision and accuracy. In the 3-text sampling scheme, the cluster bootstrap results in the smallest interval but provides insufficient coverage. The regularized cluster bootstrap is the next most precise with coverage that is close to 0.95 across the sample sizes. In the 30-text case, however, the cluster bootstrap is inappropriate, never reaching 0.95 coverage and under 0.90 when the number of participants is $<20$. Instead, the Agresti-Coull interval with $n_{\text{eff}}$ \& $\text{df}_{\text{design}}$ adjustments provides the best coverage, though with intervals that are wider than the other methods when the number of individuals is fewer than 10.

\subsection{Sample Size Considerations}

\begin{figure}[htbp]
\centering
\includegraphics[width=0.75\linewidth]{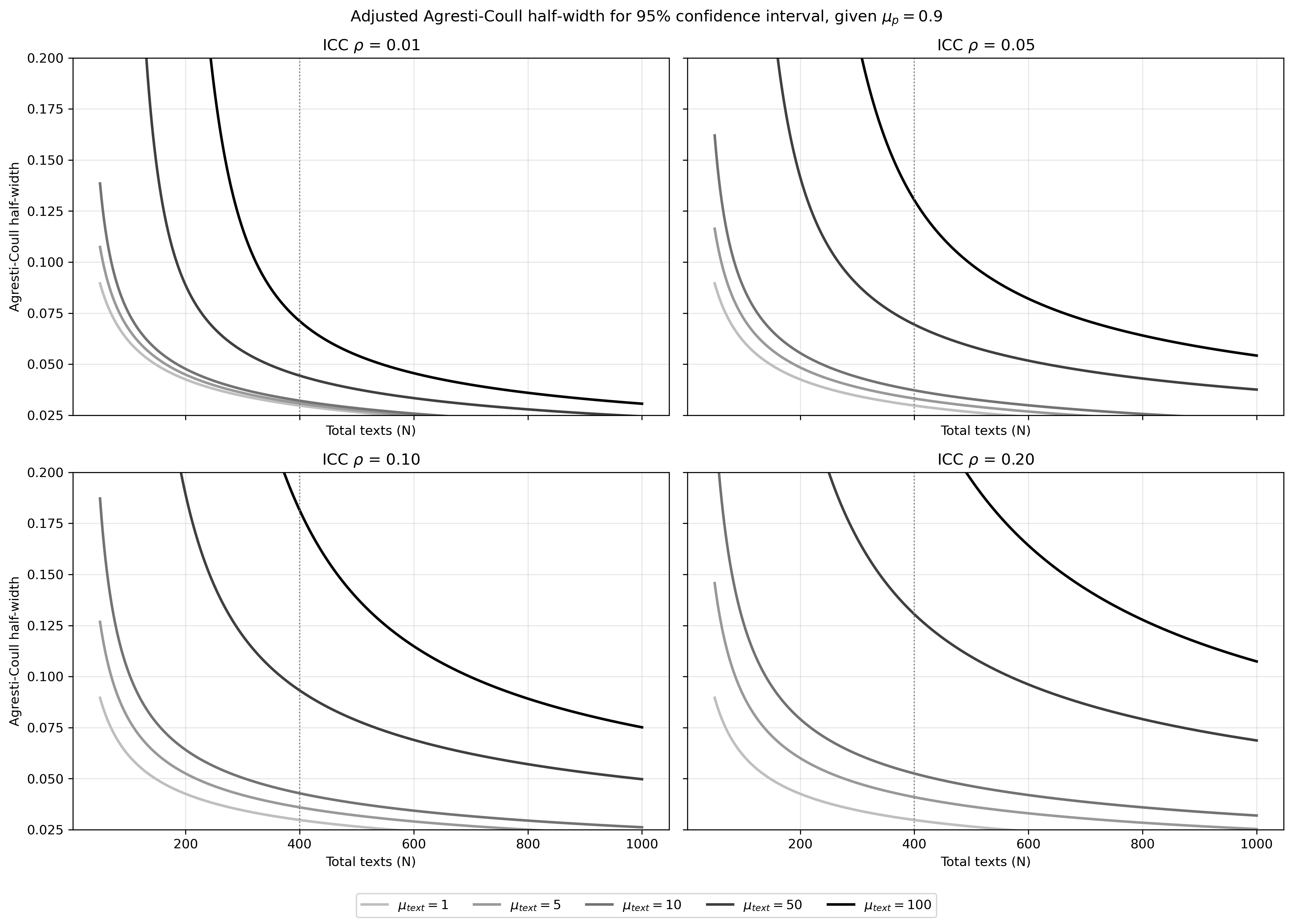}
\caption{Resulting 95\% Confidence Interval Half-Width by Sample Size at
Metric = 0.90, Varying Cluster Size and ICC}
\label{fig:fig7}
\end{figure}

Figure~\ref{fig:fig7} displays the relationship between the number of texts used to estimate performance and the half width of an adjusted 95\% Agresti-Coull confidence interval, holding the performance rate at 0.90. As expected, confidence interval width increases with ICC values, and the importance of ICC increases with the number of texts per cluster. Holding the number of texts constant, fewer texts per cluster can dramatically reduce confidence interval width. Consider an ICC of $\rho = 0.1$ and a situation where the analyst has the resources to hand-label 400 texts for the calculation of a metric. (Again, in the context of validating binary text classifiers, the analytic sample size is the number of observations used to estimate the metric. For recall, this would be the number of positive human classifications.) Holding the number of texts constant, at an average cluster size of 10, the half width of 95\% Agresti-Coull confidence interval would be 0.043. At an average cluster size of 50, it would be 0.093, illustrating that, whenever possible, increasing the number of individuals is preferable from the standpoint of statistical precision to increasing the number of texts within individuals.

\noindent {\footnotesize \textit{Note.} No ICC adjustment incorporated with $\mu_{\text{text}} = 1$.}

\section{Discussion}

Performance metrics are the dominant form of evidence researchers provide when demonstrating that their classifier is good enough to be of use \citep{james_introduction_2013}. Yet, unlike other forms of statistical evidence in psychology and the social sciences, these metrics are only inconsistently reported alongside estimates of uncertainty \citep{anglin_addressing_2024}. A failure to report confidence intervals reduces the transparency of the field, making it unclear how much confidence can be placed in estimated performance. This is particularly concerning when the relevant sample size---the number of individuals and texts used in the estimation of the metric---also goes unreported. Thus, a primary aim of this article is to encourage the accurate reporting of confidence intervals surrounding performance metrics and to increase attention to sample size requirements earlier in the design process.

This article makes five key contributions to that end. First, we build on prior related simulation studies \citep[e.g.,][]{agresti_approximate_1998, kahouadji_comprehensive_2025} by providing a systematic comparison of confidence interval coverage under the conditions that commonly characterize text classification in the social sciences: high performance levels (increasingly expected with LLMs; \citealp{gilardi_chatgpt_2023}), small to moderate sample sizes, infrequent constructs, and texts nested within individuals. Second, we bring attention to the limitations of the percentile bootstrap for intervals surrounding proportions, building on related literature not tailored to conditions of text classification \citep[e.g.,][]{pires_interval_2008}. Third, we propose a simple pseudo-count regularization of the bootstrap that improves coverage at small sample sizes and high performance values. Fourth, we provide straightforward recommendations for nested data, illustrating the requirements for achieving accurate coverage. Finally, the results here highlight the importance of obtaining a sufficient \emph{relevant} sample size: a sufficient number of observations used to calculate the metric (for recall, the number of positive human classifications; for precision the number of positive machine classifications) and, when texts are nested, the number of individuals rather than the number of texts. By illustrating confidence interval width under varying conditions, we aim to direct researchers' attention to sample size earlier in the study design.

Our simulations offer two key warnings. First, even in an unnested setting, Wald intervals and percentile bootstrap intervals can dramatically underestimate uncertainty. For example, at true recall or precision of 0.95 and a relevant sample size of 20, Wald interval and percentile bootstrap coverage were both estimated to be 0.66. When the relevant sample size is 10, Wald interval coverage was 0.401 and the basic percentile bootstrap coverage was 0.381. On the other hand, Agresti-Coull, Wilson, Clopper-Pearson, and pseudo-count regularized intervals all provide accurate coverage even when the relevant sample size is as small as 10. Though these very small sample sizes may appear to be a strawman, irrelevant in practice, we argue that they are not rare enough to be discounted. Because constructs are often rare, and the relevant sample size for recall/precision is the number of positive machine/human classifications, it is possible to have very few texts included in the metric calculation. Indeed, this is one reason why it is important for confidence intervals to become standard reporting criteria: to make transparent where sample sizes are limited. Further, in the best case scenario, performance metrics are not only reported for the average interval, but also for relevant subgroups---for example, distinguished by race and gender \citep{buolamwini_gender_2018}. Doing so allows for assessments of fairness and disparate impact but also requires planning for sample sizes to be large enough to support sufficiently precise inference---something that accurate confidence intervals make clear.

Second, when the texts are nested within individuals---as is common given a single person may contribute many utterances \citep[e.g.,][]{hammerfald_leveraging_2025} or writing samples \citep[e.g.,][]{jamil_monitoring_2017}---methods that ignore clustering, can substantially overstate confidence. For example, when there are between 5 and 50 individuals producing about 30 texts each, the true metric is 0.90, and ICC is approximately 0.11, the unadjusted Agresti-Coull interval (appropriate when texts are not nested), covered the true metric only about 65\% of the time. Adjusted analytic intervals incorporating adjustments for $n_{\text{eff}}$ \& $\text{df}_{\text{design}}$ address this issue, as does the pseudo-regularized hierarchical bootstrap.

Based on our simulations, we offer the following recommendations about which interval to use, and when. For independent (unnested) data, we recommend the Agresti-Coull interval as a default as it is simple to calculate and provides coverage at or near the nominal level across the performance levels and sample sizes. Wilson and Clopper-Pearson intervals are perfectly reasonable alternatives. Analysts who prefer non-analytic intervals (or who require them for non-proportion based metrics) should use a pseudo-count--regularized bootstrap rather than the basic percentile bootstrap. When texts are nested within individuals but the number of texts produced by individuals is small ($\sim$3), the pseudo-regularized cluster bootstrap is preferable, reaching sufficient coverage while also minimizing interval width. However, when the number of texts is larger, analysts should employ $n_{\text{eff}}$ \& $\text{df}_{\text{design}}$ adjusted intervals where possible, and pseudo-count regularized hierarchical intervals when non-analytic intervals are useful (as in the calculation of F1).

These recommendations come with the following limitations. First, while the non-nested simulations span conditions typical to text classification in psychology, they are not exhaustive. More pointedly, our nested simulations assume a single ICC structure and performance; real corpora will reflect more variable conditions. Second, this article does not offer any information on commonly \emph{observed} ICC values in psychology. For analysts to pro-actively plan for necessary sample sizes, benchmarks like those provided for nested classroom settings and test outcomes \citep{hedges_intraclass_2007}, will be useful. Similarly, we only anecdotally speak to common performance metric ranges; empirical evidence on commonly achieved performance values will be similarly useful for planning. Finally, our recommendations address the estimation of uncertainty surrounding the performance metrics, but only briefly touch on the unbiased estimation of performance metrics themselves, which bring issues of their own \citep[e.g., see][]{kapoor_leakage_2023}. 

\section{Conclusion}

If the magnitude of a performance metric matters, so too does its precision. Performance metrics are subject to the same sampling variation as any other quantity researchers report. Confidence intervals make this sampling variation transparent, and their absence threatens to produce overconfidence in model-based inferences. Yet, as we have shown, default approaches to confidence interval estimation are often inappropriate under conditions common to text classification in the social sciences, particularly with large language models (small labelled datasets, rare constructs, high performance, and nested data). This article provides evidence-based recommendations for accurate confidence interval reporting: Agresti-Coull or appropriate analytic alternatives for independent texts, effective sample size and degrees of freedom adjusted intervals for nested texts, and pseudo-count regularized (hierarchical) bootstrap intervals when analytic intervals are infeasible.

\printbibliography

@article{kapoor_leakage_2023,
	title = {Leakage and the reproducibility crisis in machine-learning-based science},
	volume = {4},
	doi = {https://doi.org/10.1016/j.patter.2023.100804},
	number = {9},
	journal = {Patterns},
	publisher = {Elsevier},
	author = {Kapoor, Sayash and Narayanan, Arvind},
	year = {2023},
	note = {ISBN: 2666-3899},
}

@article{cronbach_construct_1955,
	title = {Construct validity in psychological tests.},
	volume = {52},
	issn = {1939-1455},
	url = {http://psycnet.apa.org/fulltext/1956-03730-001.pdf},
	doi = {10.1037/h0040957},
	number = {4},
	urldate = {2020-05-22},
	journal = {Psychological Bulletin},
	author = {Cronbach, Lee J. and Meehl, Paul E.},
	year = {1955},
	note = {Publisher: US: American Psychological Association},
	pages = {281},
}

@inproceedings{mitchell_model_2019,
	title = {Model cards for model reporting},
	booktitle = {Proceedings of the {Conference} on {Fairness}, {Accountability}, and {Transparency}},
	url = {https://dx.doi.org/10.1145/3287560.3287596},
	doi = {10.1145/3287560.3287596},
	publisher = {ACM},
	author = {Mitchell, Margaret and Wu, Simone and Zaldivar, Andrew and Barnes, Parker and Vasserman, Lucy and Hutchinson, Ben and Spitzer, Elena and Raji, Inioluwa Deborah and Gebru, Timnit},
	year = {2019},
}

@book{cohen_statistical_1988,
	title = {Statistical power analysis for the behavioral sciences},
	isbn = {0-203-77158-3},
	publisher = {Routledge},
	author = {Cohen, Jacob},
	year = {1988},
}

@article{hedges_statistical_2000,
	title = {Statistical power analysis in education research ({NCSER} 2010-3006)},
	volume = {1},
	url = {http://ies.ed.gov/ncser/},
	doi = {10.1037/e599842011-001},
	journal = {National Center for Special Education Research},
	author = {Hedges, Larry V and Rhoads, C},
	year = {2000},
	pages = {88--88},
}

@inproceedings{shaffer_how_2021,
	address = {Malibu, CA},
	title = {How {We} {Code}},
	booktitle = {{ICQE} 2020},
	publisher = {Springer},
	author = {Shaffer, David Williamson and Ruis, Andrew R.},
	editor = {Lee, S.B. and {A.R. Ruis}},
	month = jan,
	year = {2021},
}

@article{hedges_intraclass_2007,
	title = {Intraclass correlation values for planning group-randomized trials in education},
	volume = {29},
	doi = {https://doi.org/10.3102/0162373707299},
	number = {1},
	journal = {Educational Evaluation and Policy Analysis},
	author = {Hedges, Larry V and Hedberg, E. C.},
	year = {2007},
	pages = {60--87},
}

@book{grimmer_text_2022,
	title = {Text as data: {A} new framework for machine learning and the social sciences},
	isbn = {0-691-20799-2},
	publisher = {Princeton University Press},
	author = {Grimmer, Justin and Roberts, Margaret E. and Stewart, Brandon M.},
	year = {2022},
}

@inproceedings{green_algorithmic_2020,
	title = {Algorithmic realism: expanding the boundaries of algorithmic thought},
	url = {https://doi.org/10.1145/3351095.3372840},
	booktitle = {Proceedings of the 2020 conference on fairness, accountability, and transparency},
	author = {Green, Ben and Viljoen, Salomé},
	year = {2020},
	pages = {19--31},
}

@inproceedings{buolamwini_gender_2018,
	title = {Gender shades: {Intersectional} accuracy disparities in commercial gender classification},
	isbn = {2640-3498},
	url = {https://proceedings.mlr.press/v81/buolamwini18a.html},
	booktitle = {Conference on fairness, accountability and transparency},
	publisher = {PMLR},
	author = {Buolamwini, Joy and Gebru, Timnit},
	year = {2018},
	pages = {77--91},
}

@book{james_introduction_2013,
	edition = {2nd},
	title = {An {Introduction} to {Statistical} {Learning}},
	isbn = {978-1-0716-1417-4},
	url = {https://link.springer.com/book/10.1007/978-1-0716-1418-1},
	publisher = {Springer},
	author = {James, Gareth and Witten, Daniela and Hastie, Trevor and Tibshirani, Robert},
	year = {2013},
}

@article{stavropoulos_shadows_2024,
	title = {Shadows of wisdom: {Classifying} meta-cognitive and morally grounded narrative content via large language models},
	doi = {https://doi.org/10.3758/s13428-024-02441-0},
	journal = {Behavior Research Methods},
	publisher = {Springer},
	author = {Stavropoulos, Alexander and Crone, Damien L. and Grossmann, Igor},
	year = {2024},
	note = {ISBN: 1554-3528},
	pages = {1--15},
}

@article{gilardi_chatgpt_2023,
	title = {{ChatGPT} outperforms crowd workers for text-annotation tasks},
	volume = {120},
	doi = {https://doi.org/10.1073/pnas.230501612},
	number = {30},
	journal = {Proceedings of the National Academy of Sciences},
	publisher = {National Acad Sciences},
	author = {Gilardi, Fabrizio and Alizadeh, Meysam and Kubli, Maël},
	year = {2023},
	note = {ISBN: 0027-8424},
	pages = {e2305016120},
}

@article{anglin_addressing_2024,
	title = {Addressing threats to validity in supervised machine learning: {A} framework and best practices for education researchers},
	volume = {10},
	issn = {2332-8584},
	url = {https://doi.org/10.1177/23328584241303495},
	doi = {10.1177/23328584241303495},
	urldate = {2024-12-20},
	journal = {AERA Open},
	publisher = {SAGE Publications Inc},
	author = {Anglin, Kylie L.},
	month = dec,
	year = {2024},
	pages = {1--21},
}

@book{gwet_handbook_2014,
	title = {Handbook of inter-rater reliability: {The} definitive guide to measuring the extent of agreement among raters},
	isbn = {0-9708062-8-0},
	publisher = {Advanced Analytics, LLC},
	author = {Gwet, Kilem L.},
	year = {2014},
}

@article{kahouadji_comprehensive_2025,
	title = {A {Comprehensive} {Comparison} of the {Wald}, {Wilson}, and adjusted {Wilson} {Confidence} {Intervals} for {Proportions}},
	journal = {arXiv preprint arXiv:2508.10223},
	author = {Kahouadji, Nabil},
	year = {2025},
}

@article{agresti_approximate_1998,
	title = {Approximate is better than “exact” for interval estimation of binomial proportions},
	volume = {52},
	number = {2},
	journal = {The American Statistician},
	publisher = {Taylor \& Francis},
	author = {Agresti, Alan and Coull, Brent A.},
	year = {1998},
	note = {ISBN: 0003-1305},
	pages = {119--126},
}

@book{kish_survey_1965,
	title = {Survey sampling},
	publisher = {Wiley},
	author = {Kish, Leslie},
	year = {1965},
}

@article{clopper_use_1934,
	title = {The use of confidence or fiducial limits illustrated in the case of the binomial},
	volume = {26},
	number = {4},
	journal = {Biometrika},
	publisher = {JSTOR},
	author = {Clopper, Charles J. and Pearson, Egon S.},
	year = {1934},
	note = {ISBN: 0006-3444},
	pages = {404--413},
}

@article{tibshirani_introduction_1993,
	title = {An introduction to the bootstrap},
	volume = {57},
	number = {1},
	journal = {Monographs on statistics and applied probability},
	author = {Tibshirani, Robert J. and Efron, Bradley},
	year = {1993},
	pages = {1--436},
}

@article{raschka_model_2018,
	title = {Model evaluation, model selection, and algorithm selection in machine learning},
	journal = {arXiv preprint arXiv:1811.12808},
	author = {Raschka, Sebastian},
	year = {2018},
}

@article{rainio_evaluation_2024,
	title = {Evaluation metrics and statistical tests for machine learning},
	volume = {14},
	number = {1},
	journal = {Scientific Reports},
	publisher = {Nature Publishing Group UK London},
	author = {Rainio, Oona and Teuho, Jarmo and Klén, Riku},
	year = {2024},
	note = {ISBN: 2045-2322},
	pages = {6086},
}

@article{efron_nonparametric_1981,
	title = {Nonparametric standard errors and confidence intervals},
	volume = {9},
	number = {2},
	journal = {canadian Journal of Statistics},
	publisher = {Wiley Online Library},
	author = {Efron, Bradley},
	year = {1981},
	note = {ISBN: 0319-5724},
	pages = {139--158},
}

@inproceedings{japkowicz_why_2006,
	title = {Why question machine learning evaluation methods},
	volume = {6},
	number = {11},
	booktitle = {{AAAI} workshop on evaluation methods for machine learning},
	publisher = {University of Ottawa},
	author = {Japkowicz, Nathalie},
	year = {2006},
}

@inproceedings{ling_auc_2003,
	title = {{AUC}: a better measure than accuracy in comparing learning algorithms},
	booktitle = {Conference of the canadian society for computational studies of intelligence},
	publisher = {Springer},
	author = {Ling, Charles X. and Huang, Jin and Zhang, Harry},
	year = {2003},
	pages = {329--341},
}

@article{han_determination_2022,
	title = {Determination of the number of observers needed to evaluate a subjective test and its application in two {PD}‐{L1} studies},
	volume = {41},
	number = {8},
	journal = {Statistics in medicine},
	publisher = {Wiley Online Library},
	author = {Han, Gang and Schell, Michael J. and Reisenbichler, Emily S. and Guo, Bohong and Rimm, David L.},
	year = {2022},
	note = {ISBN: 0277-6715},
	pages = {1361--1375},
}

@article{diciccio_review_1988,
	title = {A {Review} of {Bootstrap} {Confidence} {Intervals}},
	volume = {50},
	url = {https://doi.org/10.1111/j.2517-6161.1988.tb01732.x},
	number = {3},
	journal = {Journal of the Royal Statistical Society: Series B (Methodological)},
	author = {Diciccio, Thomas J. and Romano, Joseph P.},
	year = {1988},
	note = {ISBN: 0035-9246
Type: 10.1111/j.2517-6161.1988.tb01732.x},
	pages = {338--354},
}

@article{hesterberg_bootstrap_2011,
	title = {Bootstrap},
	volume = {3},
	url = {https://wires.onlinelibrary.wiley.com/doi/abs/10.1002/wics.182},
	number = {6},
	journal = {WIREs Comp Stat},
	author = {Hesterberg, Tim},
	year = {2011},
	note = {Pages: 497-526
Publication Title: WIREs Computational Statistics
Type: https://doi.org/10.1002/wics.182},
	pages = {497--526},
}

@article{dean_evaluating_2015,
	title = {Evaluating confidence interval methods for binomial proportions in clustered surveys},
	volume = {3},
	number = {4},
	journal = {Journal of Survey Statistics and Methodology},
	publisher = {Oxford University Press},
	author = {Dean, Natalie and Pagano, Marcello},
	year = {2015},
	note = {ISBN: 2325-0992},
	pages = {484--503},
}

@article{wu_comparison_2012,
	title = {Comparison of methods for estimating the intraclass correlation coefficient for binary responses in cancer prevention cluster randomized trials},
	volume = {33},
	issn = {1551-7144},
	url = {https://www.sciencedirect.com/science/article/pii/S1551714412001310},
	doi = {10.1016/j.cct.2012.05.004},
	number = {5},
	journal = {Contemporary Clinical Trials},
	author = {Wu, Sheng and Crespi, Catherine M. and Wong, Weng Kee},
	month = sep,
	year = {2012},
	pages = {869--880},
}

@article{korn_confidence_1998,
	title = {Confidence intervals for proportions with small expected number of positive counts estimated from survey data},
	volume = {24},
	journal = {Survey Methodology},
	publisher = {STATISTICS CANADA},
	author = {Korn, Edward L. and Graubard, Barry I.},
	year = {1998},
	note = {ISBN: 0714-0045},
	pages = {193--201},
}

@article{hammerfald_leveraging_2025,
	title = {Leveraging large language models to identify microcounseling skills in psychotherapy transcripts},
	doi = {https://doi.org/10.1080/10503307.2025.2539405},
	journal = {Psychotherapy Research},
	publisher = {Taylor \& Francis},
	author = {Hammerfald, Karin and Schmidt, Fabian and Vlassov, Vladimir and Haaland Jahren, Henrik and Solbakken, Ole André},
	year = {2025},
	note = {ISBN: 1050-3307},
	pages = {1--19},
}

@inproceedings{jamil_monitoring_2017,
	title = {Monitoring tweets for depression to detect at-risk users},
	url = {https://aclanthology.org/W17-3104/},
	booktitle = {Proceedings of the fourth workshop on computational linguistics and clinical psychology—from linguistic signal to clinical reality},
	author = {Jamil, Zunaira and Inkpen, Diana and Buddhitha, Prasadith and White, Kenton},
	year = {2017},
	pages = {32--40},
}

@article{saravanan_application_2020,
	title = {Application of the hierarchical bootstrap to multi-level data in neuroscience},
	volume = {3},
	number = {5},
	journal = {Neurons, behavior, data analysis and theory},
	author = {Saravanan, Varun and Berman, Gordon J. and Sober, Samuel J.},
	year = {2020},
	pages = {https://nbdt. scholasticahq. com/article/13927--application--of--the--hierarchical--bootstrap--to--multi--level--data--in--neuroscience},
}

@article{field_bootstrapping_2007,
	title = {Bootstrapping clustered data},
	volume = {69},
	number = {3},
	journal = {Journal of the Royal Statistical Society Series B: Statistical Methodology},
	publisher = {Oxford University Press},
	author = {Field, Christopher A. and Welsh, Alan H.},
	year = {2007},
	note = {ISBN: 1369-7412},
	pages = {369--390},
}

@article{mccullagh_resampling_2000,
	title = {Resampling and exchangeable arrays},
	author = {McCullagh, Peter},
	journal = {Bernoulli},
	volume = {6},
	number = {2},
	year = {2000},
}

@article{pires_interval_2008,
	title = {Interval estimators for a binomial proportion: {Comparison} of twenty methods},
	volume = {6},
	number = {2},
	journal = {REVSTAT-Statistical Journal},
	author = {Pires, Ana M. and Amado, Conceiçao},
	year = {2008},
	note = {ISBN: 2183-0371},
	pages = {165--197},
}

\clearpage
\appendix
\setcounter{figure}{0}
\counterwithin{figure}{section}
\renewcommand{\thefigure}{\thesection\arabic{figure}}

\section{The Behavior of Analytic Confidence Intervals}
\label{app:A}

\begin{figure}[htbp]
\centering
\includegraphics[width=\linewidth]{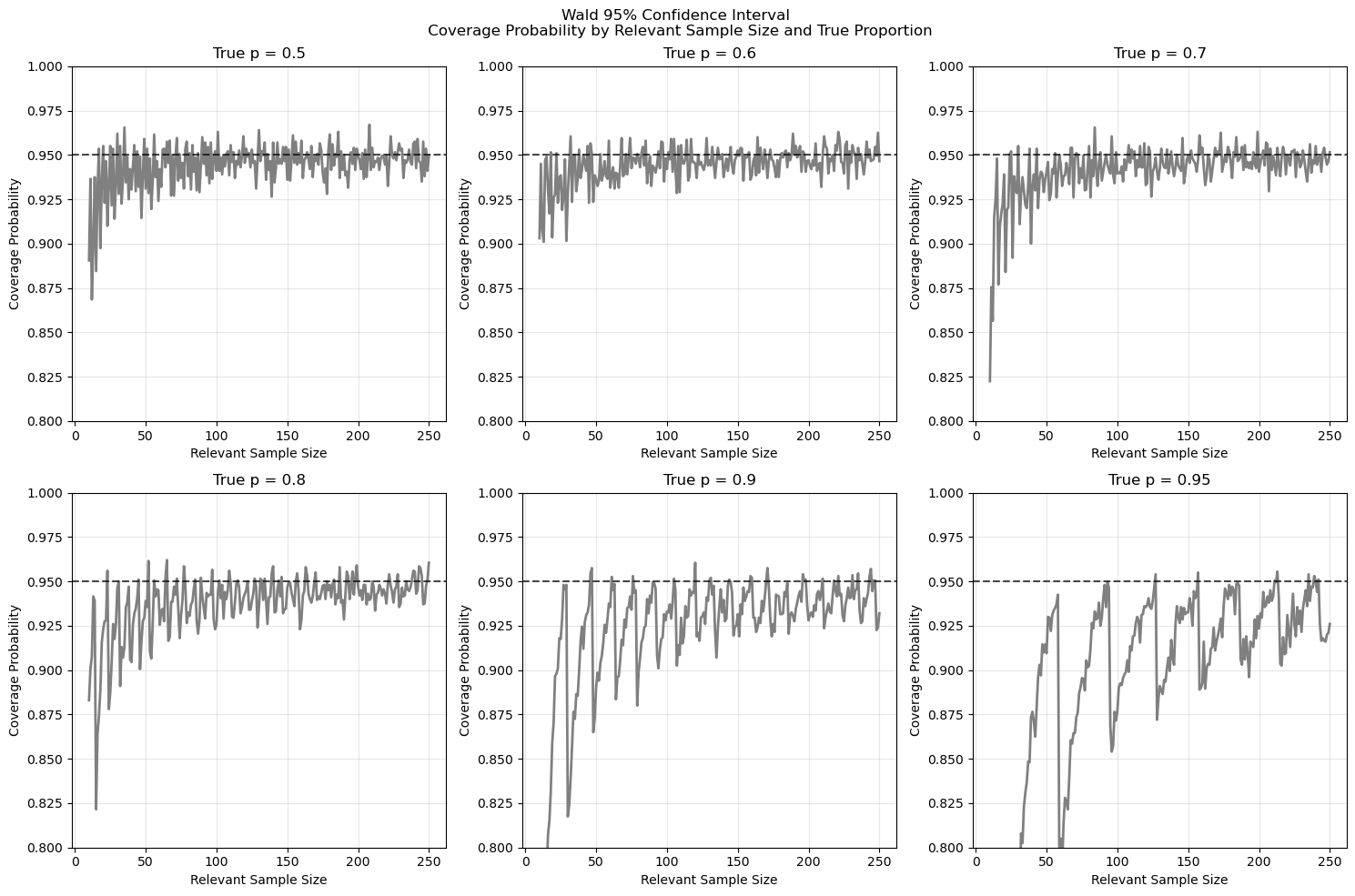}
\caption{Estimated Coverage of the Wald 95\% Confidence Interval}
\label{fig:figA1}
\end{figure}

\begin{figure}[htbp]
\centering
\includegraphics[width=\linewidth]{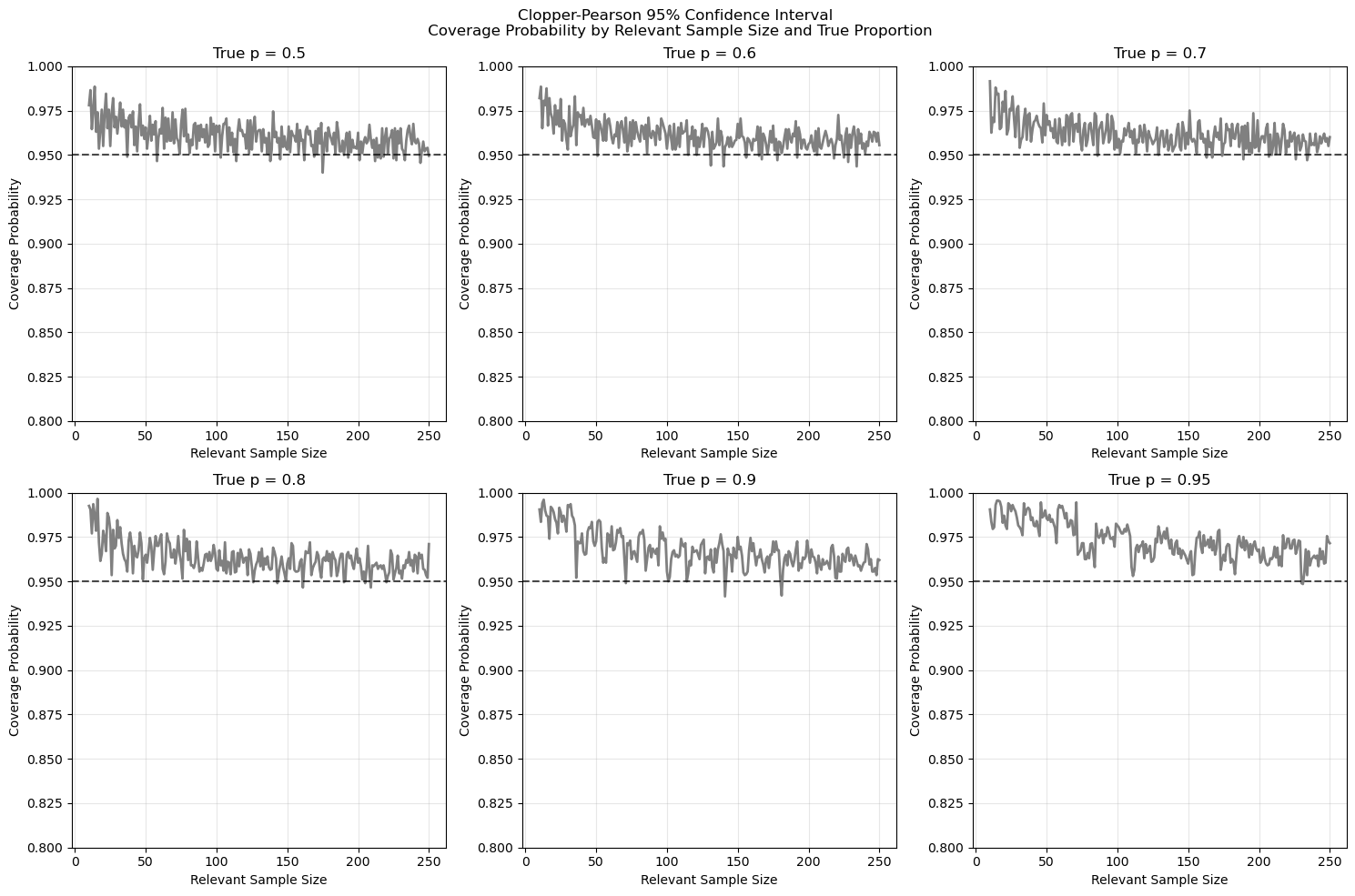}
\caption{Estimated Coverage of the Clopper-Pearson 95\% Confidence Interval}
\label{fig:figA2}
\end{figure}

\begin{figure}[htbp]
\centering
\includegraphics[width=\linewidth]{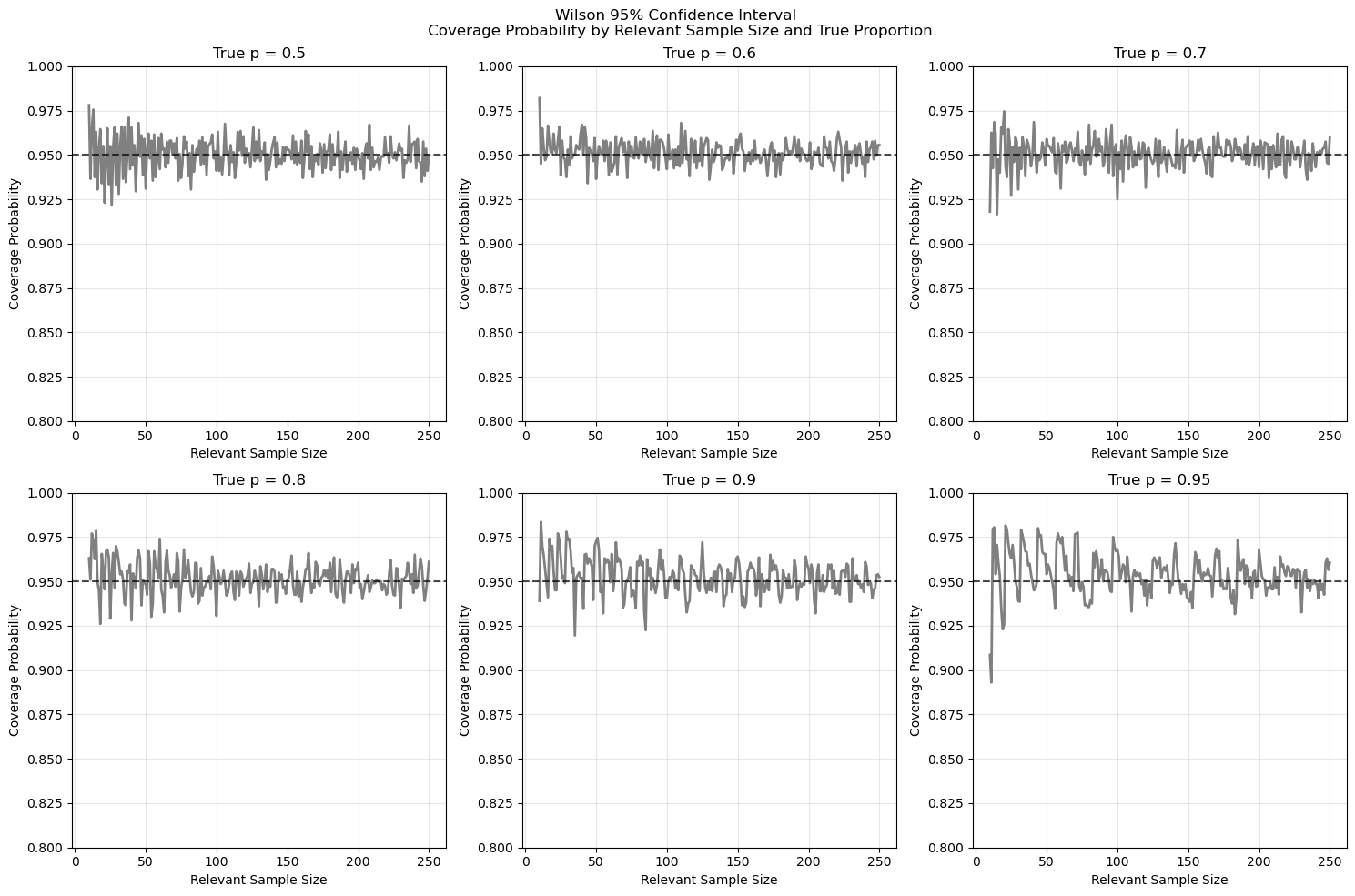}
\caption{Estimated Coverage of the Wilson 95\% Confidence Interval}
\label{fig:figA3}
\end{figure}

\begin{figure}[htbp]
\centering
\includegraphics[width=\linewidth]{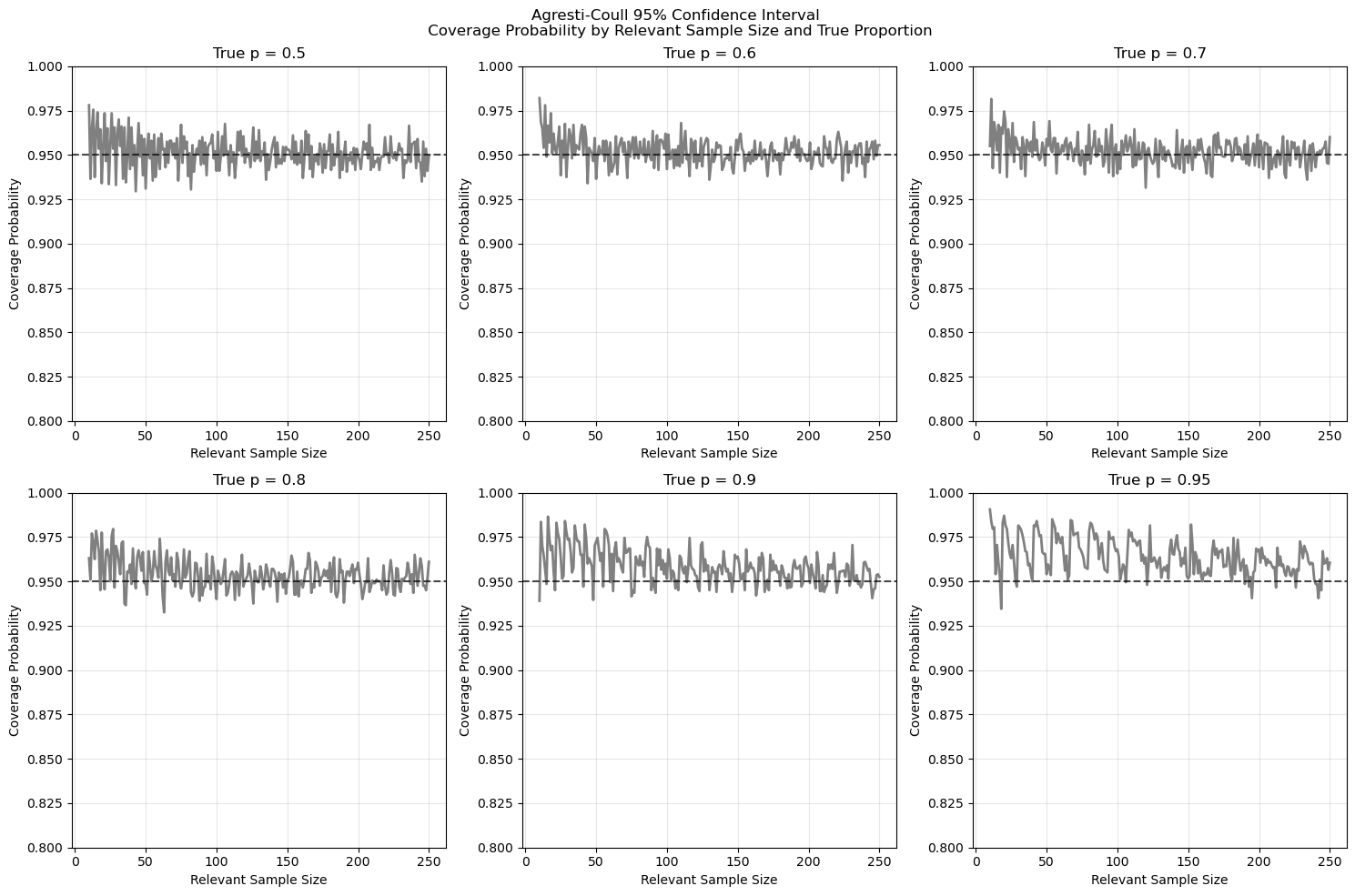}
\caption{Estimated Coverage of the Agresti-Coull 95\% Confidence Interval}
\label{fig:figA4}
\end{figure}

\begin{figure}[htbp]
\centering
\includegraphics[width=\linewidth]{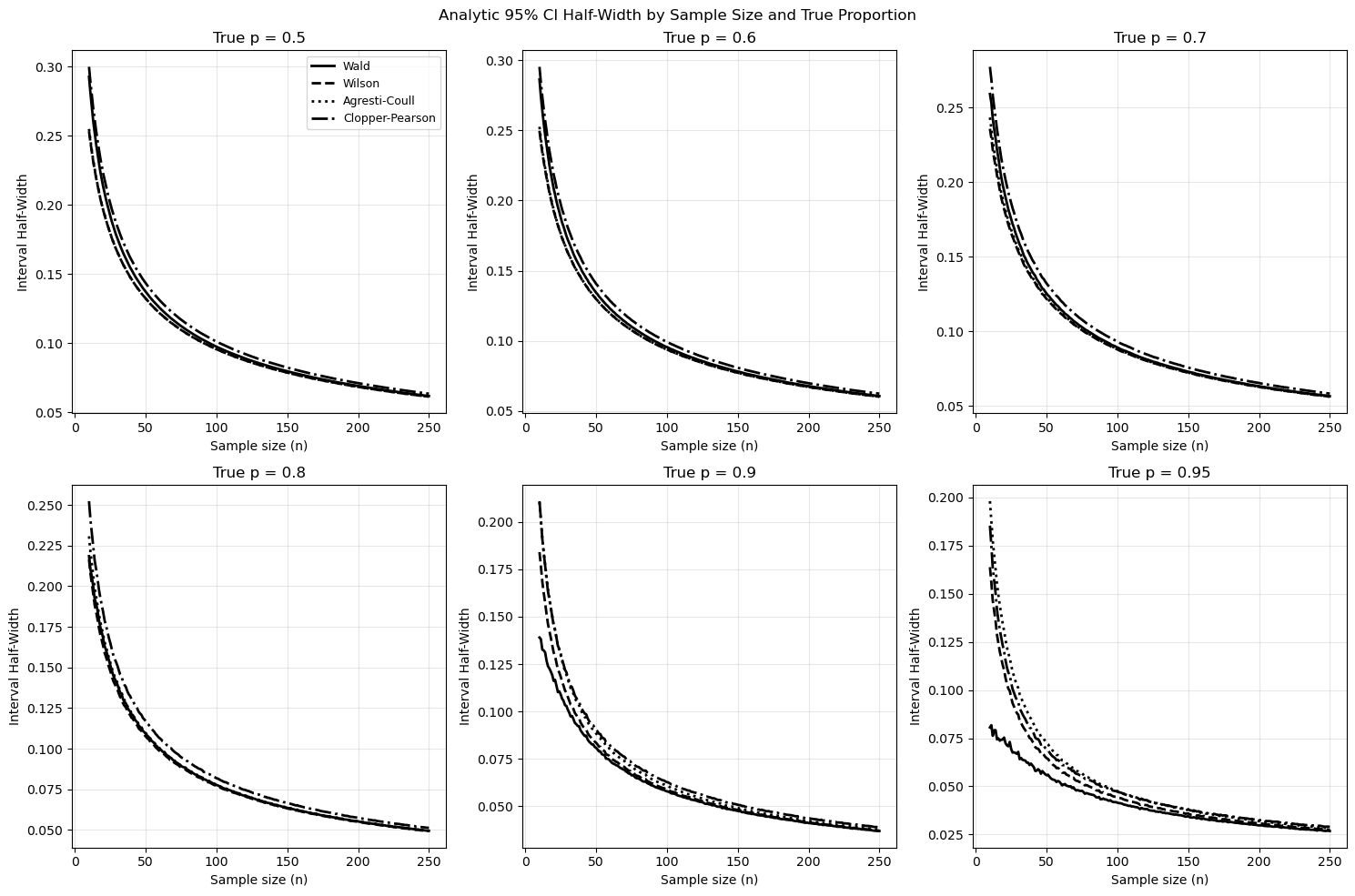}
\caption{Confidence Interval Half-Width by Analytic Method, at $n < 100$}
\label{fig:figA5}
\end{figure}

Figure~\ref{fig:figA1} confirms the findings of previous work, demonstrating the volatility and inaccuracy of the Wald interval. Across values of $p$, coverage is consistently below 0.95, with particularly low coverage rates and high volatility at low $n$ and high $p$. Unsurprisingly, Figure~\ref{fig:figA2} confirms the high coverage rates of the Clopper-Pearson standard error, which (almost) always exceeds 0.95, no matter $p$ or $n$. Comparatively, the Wilson interval (Figure~\ref{fig:figA3}), is less conservative than the Clopper-Pearson interval but still substantially more accurate and less volatile than the Wald Interval. At their lowest, Wilson coverage rates reach only 0.89, at a sample size of 11 and a $p$ of 0.95, an extreme combination. Most commonly Wilson interval coverage rates are within 0.025 of the 0.95 we expect. The Agresti-Coull performs similarly, albeit more conservatively, with coverage never reaching below 0.925 and routinely above 0.95. Figure~\ref{fig:figA5} plots the confidence interval half-width for the Wilson, Agresti-Coull and Clopper-Pearson methods. These demonstrate the modest cost of using Clopper-Pearson intervals. Taking these figures together, excluding the Wald interval, the remaining analytic options are all reasonable and can perhaps be left to the user's preference. Here, we prefer the Agresti-Coull for its high rates of coverage and its intuitiveness.

\section{The Behavior of Bootstrapped Confidence Intervals}
\label{app:B}

\begin{figure}[htbp]
\centering
\includegraphics[width=\linewidth]{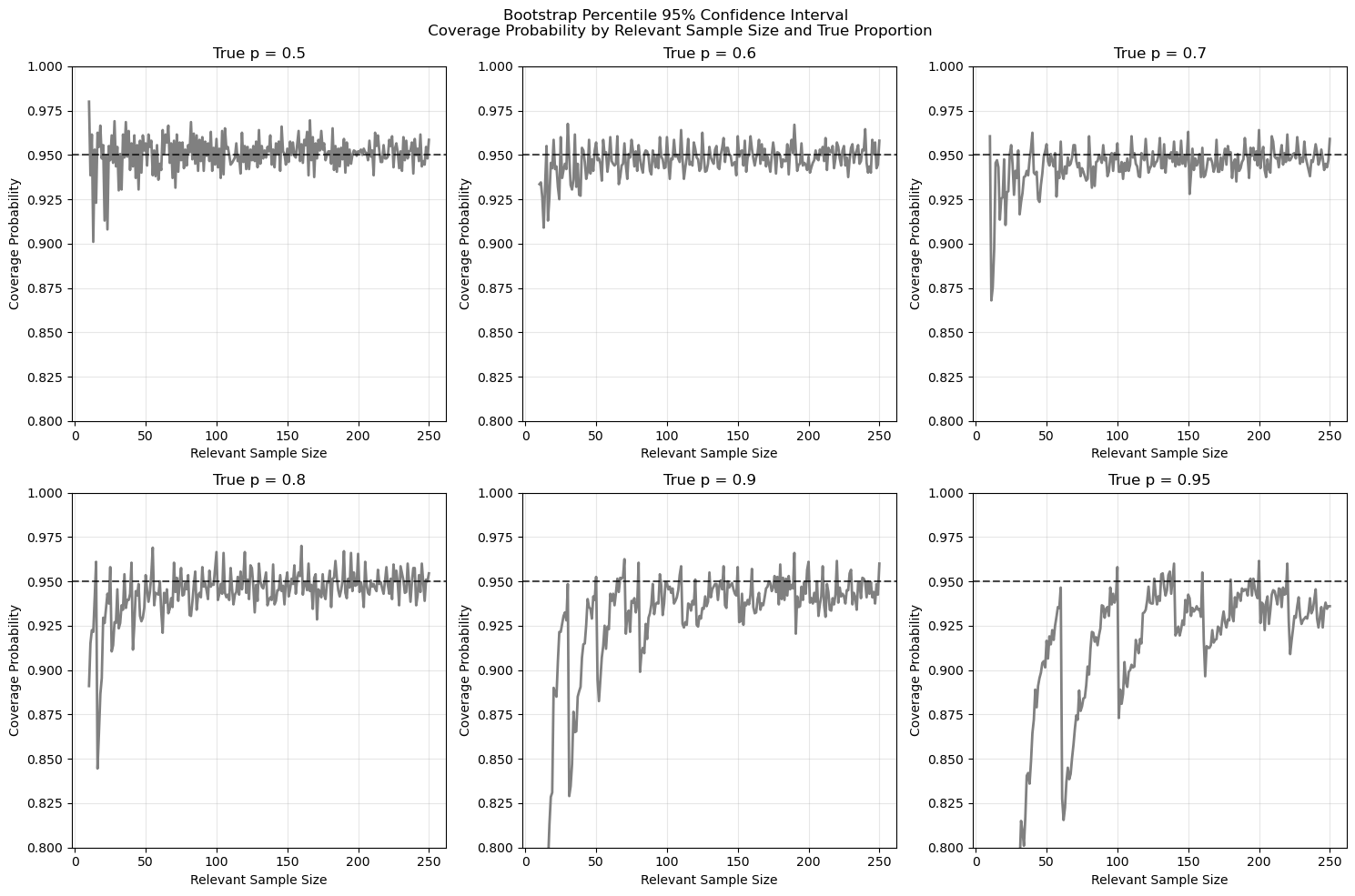}
\caption{Estimated Coverage of the Percentile Bootstrap 95\% Confidence
Interval}
\label{fig:figB1}
\end{figure}

\begin{figure}[htbp]
\centering
\includegraphics[width=\linewidth]{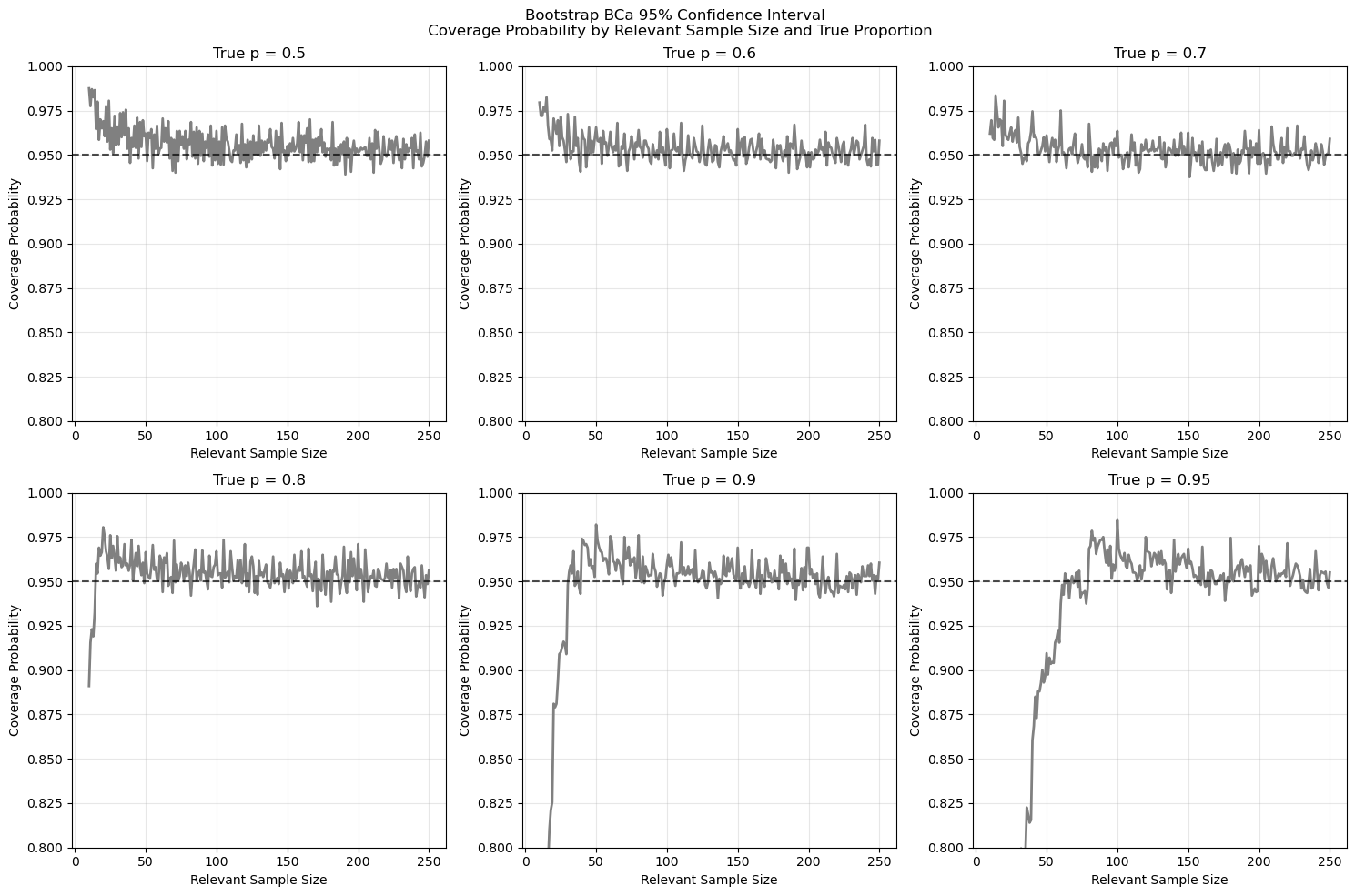}
\caption{Estimated Coverage of the BCa Bootstrap 95\% Confidence Interval}
\label{fig:figB2}
\end{figure}

\begin{figure}[htbp]
\centering
\includegraphics[width=\linewidth]{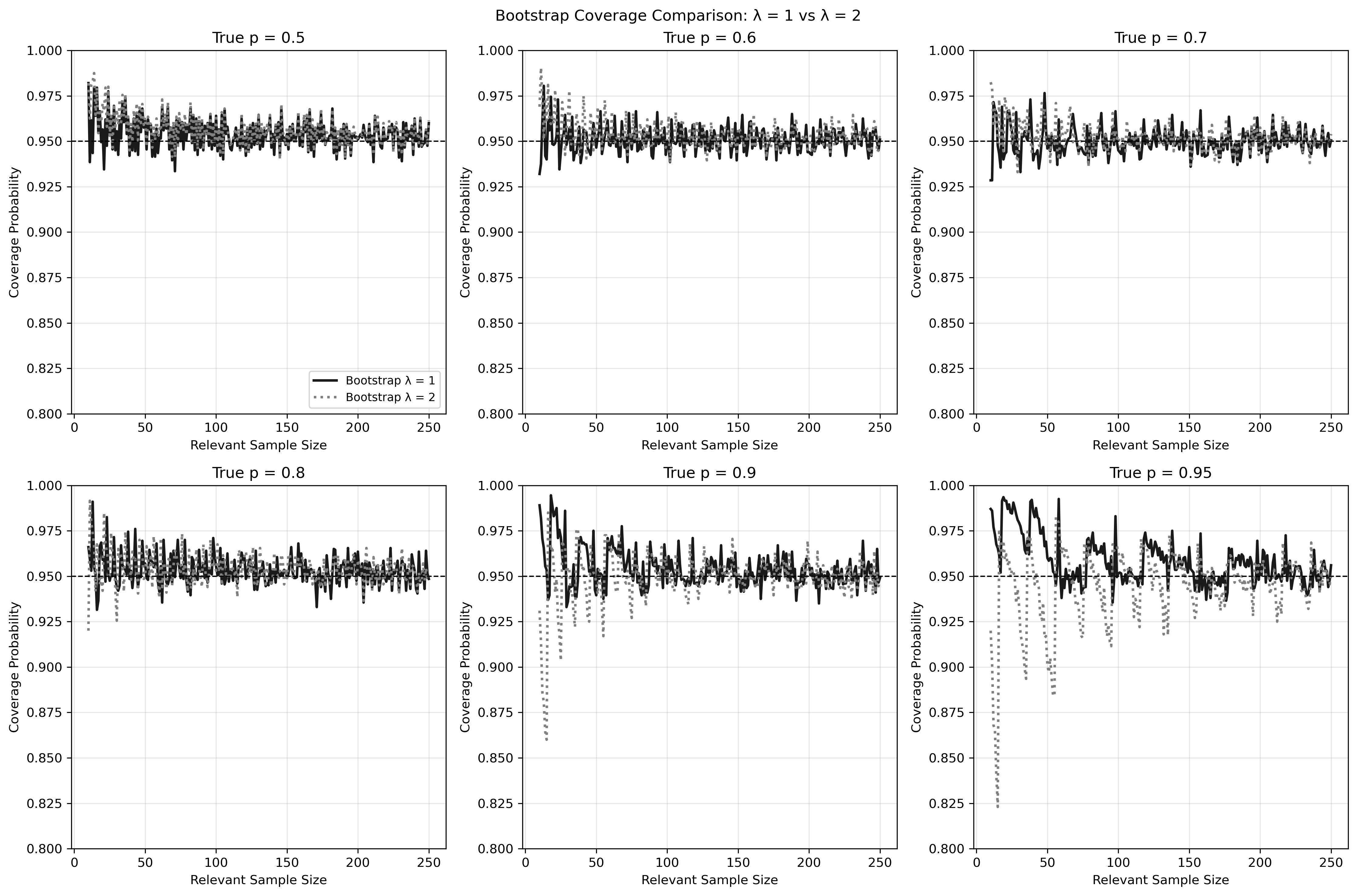}
\caption{Estimated Coverage of the Percentile Bootstrap with Pseudo Counts,
$\lambda = 1$}
\label{fig:figB3}
\end{figure}

Figure~\ref{fig:figB1} shows substantial problems with the basic bootstrapped confidence interval. When $p \geq 0.8$ and $n < 50$, coverage is routinely below 0.95 and is quite erratic. At $p = 0.9$ and $p = 0.95$, the figures show behavior that is as erratic and inaccurate as the Wald interval. The BCa interval improves upon the basic percentile interval. So long as $p < 0.9$ and $n > 50$, coverage becomes reliably above or at 0.95. However, coverage remains problematic at $n < 50$ when $p = 0.9$ and $n < 75$ at $p = 0.95$. Figure~\ref{fig:figB3} displays coverage for a pseudo-count regularized bootstraps with $\lambda = 1$ and $\lambda = 2$ (i.e., with one success/one failure and with two successes/two failures augmented to the sample before bootstrapping). For either pseudo-count approach, coverage is consistently above 0.95 until $p = 0.9$ and $p = 0.95$, at which point coverage with $\lambda = 2$ sometimes dips as low as 0.25 when $n < 50$. When $\lambda = 1$, coverage is acceptable at any sample size and any $p$.

\begin{figure}[htbp]
\centering
\includegraphics[width=\linewidth]{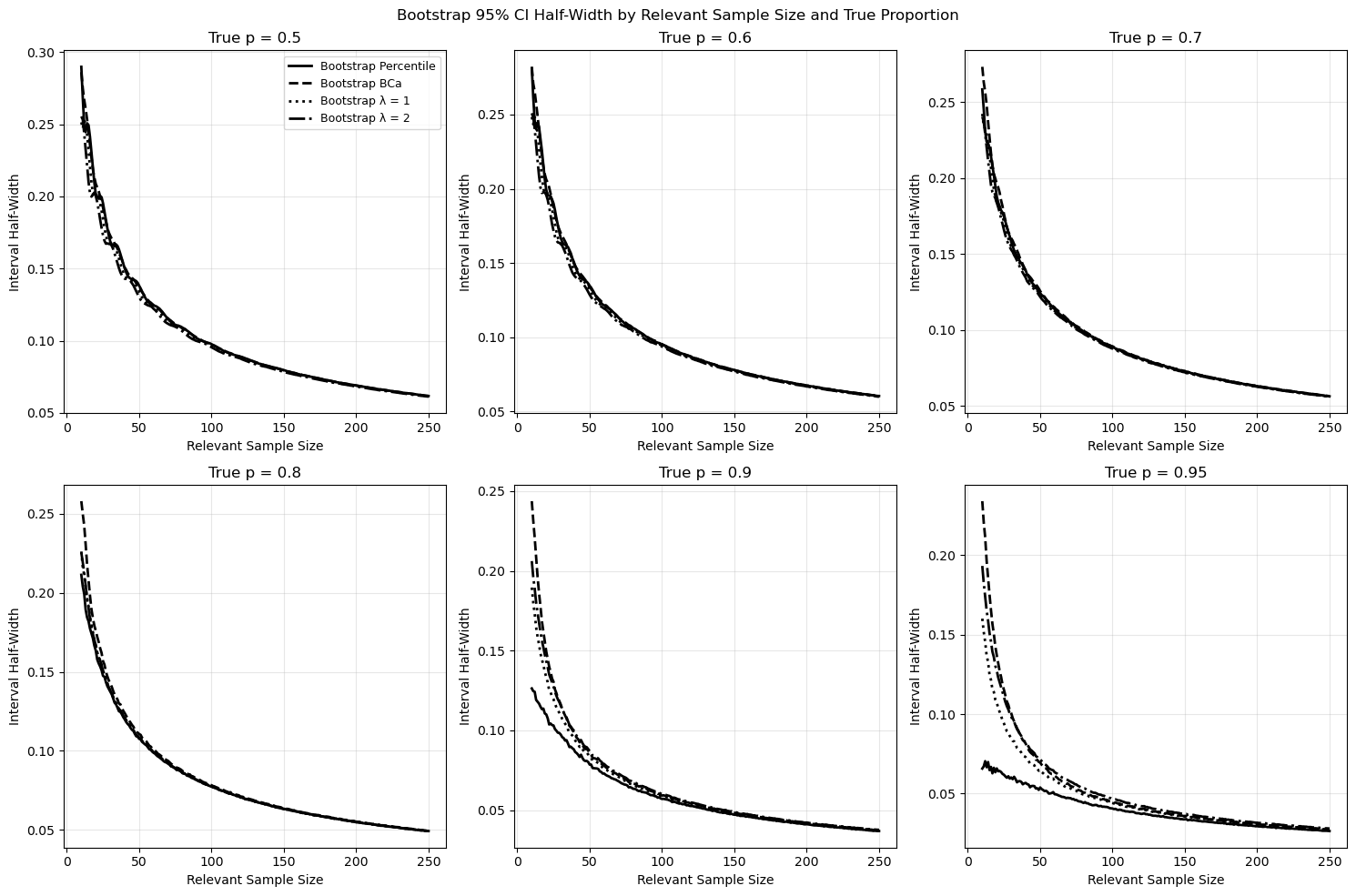}
\caption{Confidence Interval Half-Width by Bootstrapping Method, at $n < 100$}
\label{fig:figB4}
\end{figure}

Given these results, the pseudo-count regularized bootstrap with $\lambda = 1$ (one pseudo-success, one pseudo-failure) is our preferred non-parametric interval estimator. Figure~\ref{fig:figB4} demonstrates that compared to the basic percentile bootstrap it is associated with appropriately wider interval widths at high $p$ and low $n$, but produces tighter intervals than BCa.

\subsection{Variation in Bootstrap Behavior According to the Number of
Pseudo-Observations}

\begin{figure}[htbp]
\centering
\includegraphics[width=0.7\linewidth]{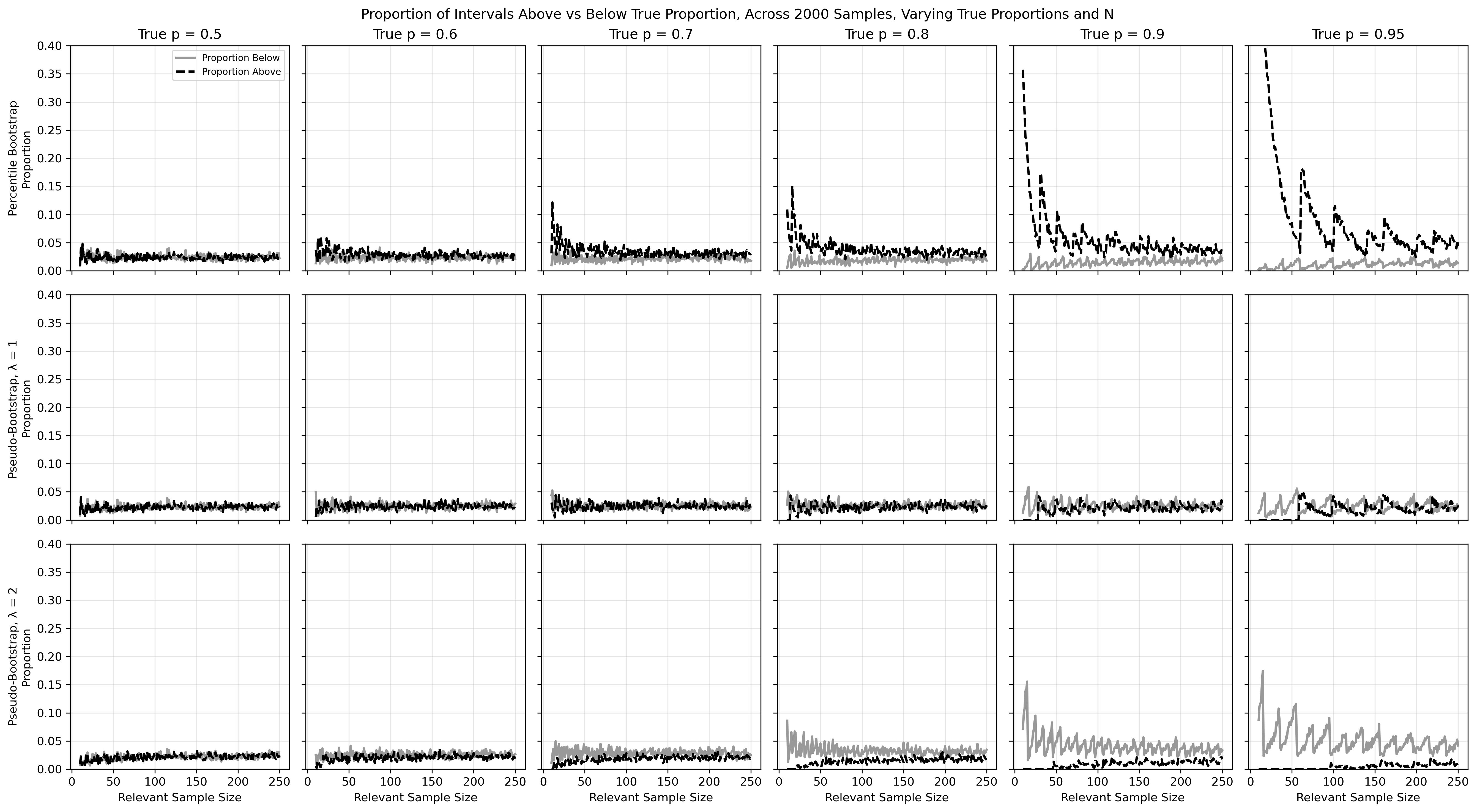}
\caption{Assessing Direction of Interval Error by Number of Pseudo-Observations,
True Proportion, and Sample Size}
\label{fig:figB5}
\end{figure}

Figure~\ref{fig:figB5} explores the direction of error in cases where a bootstrapped interval does not contain the true $p$ for a given sample. Where the basic percentile bootstrap does not contain the true $p$, this is most often because the lower bound of the interval is above the true proportion (when considering proportions greater than or equal to 0.5, as is common in the case of performance metrics). This indicates that the basic percentile bootstrap is overly optimistic, particularly when the relevant sample size is less than 50 or the metric is higher than 0.8.

On the other hand, when $\lambda = 2$, errors often occur because the \emph{upper} bound of the interval does not contain the true proportion, over-correcting for the problems with the basic bootstrap. When $\lambda = 1$, errors are roughly equally distributed in either direction, except at small sample sizes and $p = 0.95$, where errors tend to be in the pessimistic direction but still occur less than 5\% of the time.

\section{Coverage and Confidence Interval Widths under Nested Data}
\label{app:C}

\subsection{Stratified Sampling}

\begin{figure}[htbp]
\centering
\includegraphics[width=0.72\linewidth]{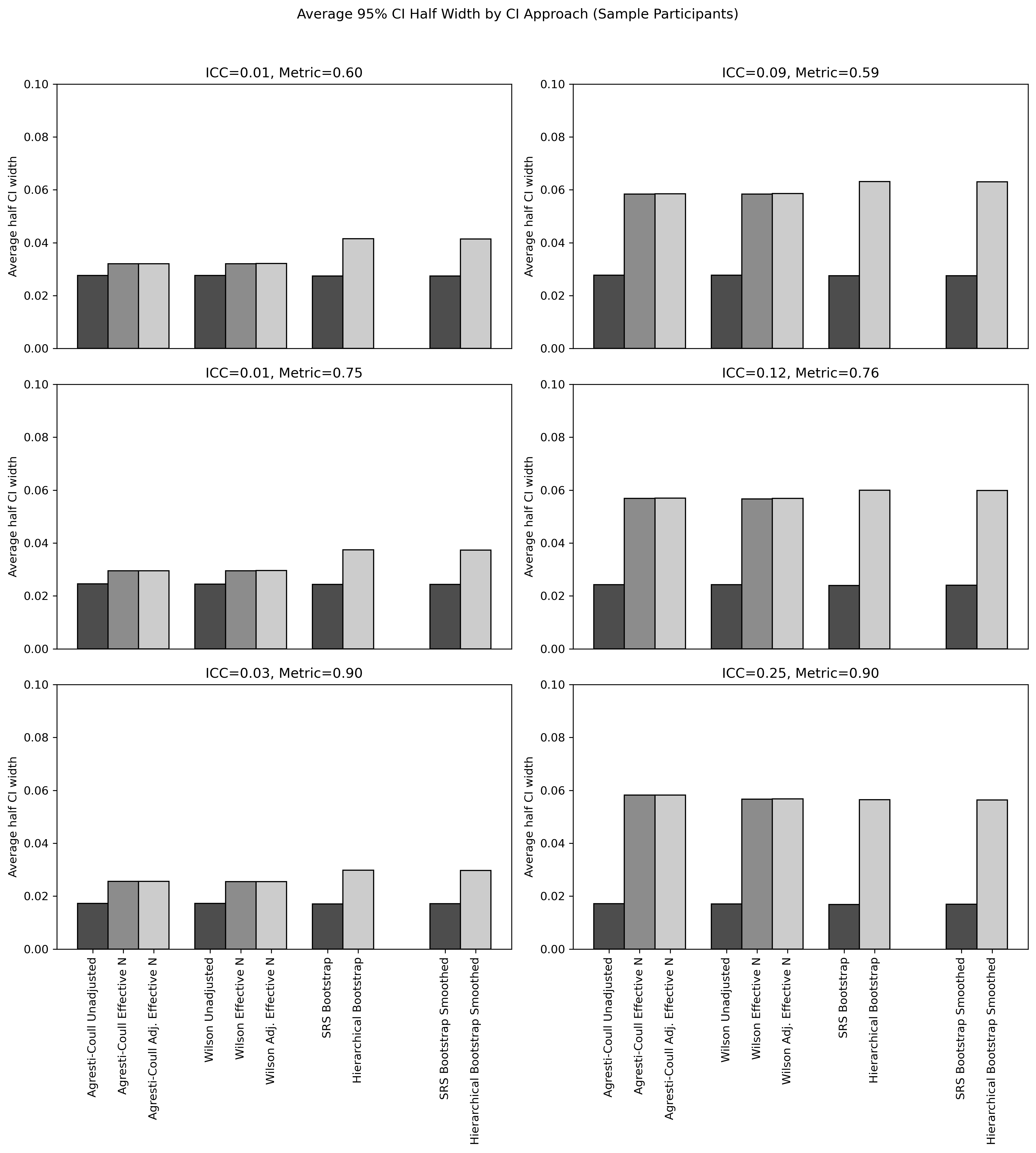}
\caption{Average Confidence Interval Width Under a Stratified Sample and Six
Simulated Populations}
\label{fig:figC1}
\end{figure}

Figure~\ref{fig:figC1} provides the average confidence interval width under each of six simulated populations (described in the main manuscript) with the validation data produced by a stratified sample of texts within individuals. Here, we see that when ICC is low (below 0.09), the adjusted analytic confidence intervals have a very small advantage in terms of precision compared to the hierarchical bootstrapped intervals, such that the half-width of the hierarchical bootstrapped intervals is approximately 0.02 wider.

\subsection{Simple Random Sample}

\begin{figure}[htbp]
\centering
\includegraphics[width=\linewidth]{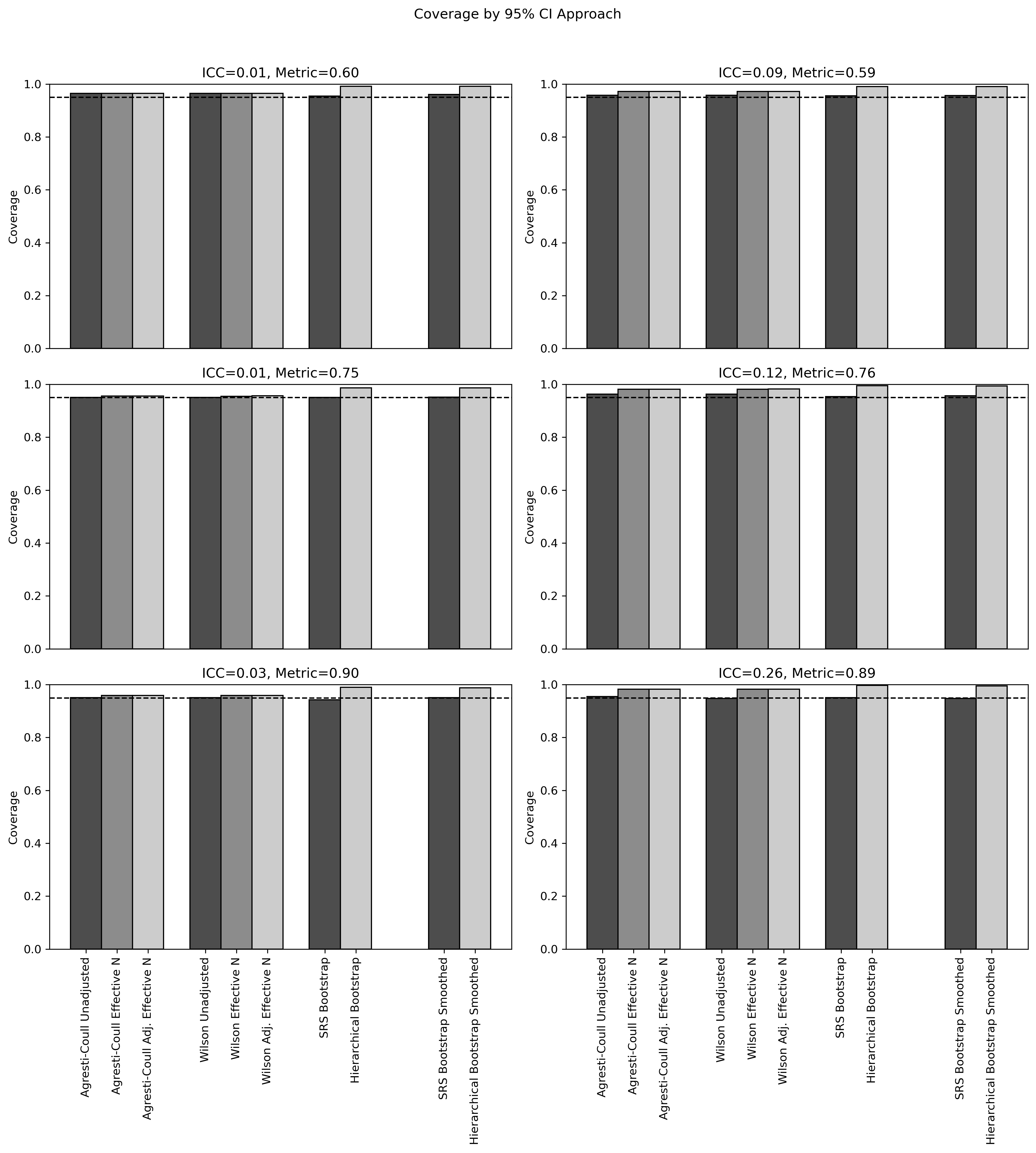}
\caption{Average 95\% Confidence Interval Coverage within a Simple Random
Sample, Under Six Simulated Populations}
\label{fig:figC2}
\end{figure}

Figure~\ref{fig:figC2} presents coverage estimates for samples of 1200 texts directly from the population, producing a sample with approximately 2.6 texts per individual. Under this scenario, in a range of ICCs, the coverage of the unadjusted confidence intervals is sufficient: approximately 0.95 across all families.

\section{Intra-Class Correlation Simulations}
\label{app:D}

\begin{figure}[htbp]
\centering
\includegraphics[width=\linewidth]{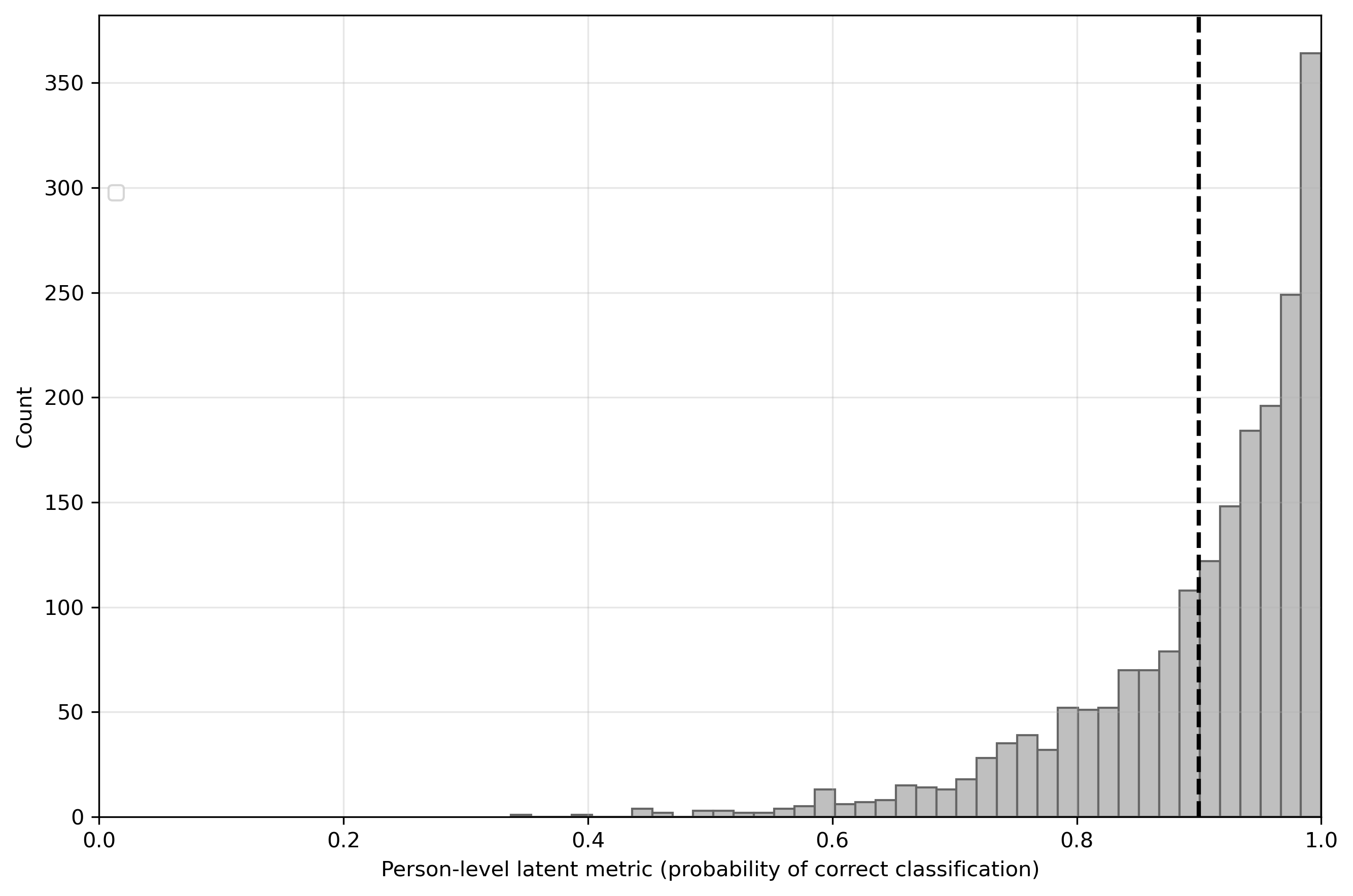}
\caption{Histogram of Simulated Finite Population,
$\mu = 0.90, \sigma = 0.10$}
\label{fig:figD1}
\end{figure}

\end{document}